\documentclass[journal]{IEEEtran}
\usepackage{xr}
\makeatletter
\newcommand*{\addFileDependency}[1]{
  \typeout{(#1)}
  \@addtofilelist{#1}
  \IfFileExists{#1}{}{\typeout{No file #1.}}
}
\makeatother

\newcommand*{\myexternaldocument}[1]{
    \externaldocument{#1}
    \addFileDependency{#1.tex}
    \addFileDependency{#1.aux}
}

\myexternaldocument{supplement}

\usepackage[version=3]{mhchem} 
\usepackage{textcomp}
\usepackage{amsmath}
\usepackage{gensymb}
\usepackage{graphicx}
\usepackage{mwe}
\usepackage{subfig}
\usepackage{color}
\usepackage{xcolor}
\usepackage{amsfonts}
\usepackage{booktabs}
\usepackage{siunitx}
\usepackage{multirow}
\usepackage{dirtytalk}
\usepackage{algpseudocode}
\usepackage{algorithm2e}
\usepackage{graphicx}
\usepackage{epstopdf}
\usepackage{tikz}
\usepackage{anyfontsize}
\usepackage{tabularx}
\usepackage[noadjust]{cite}
\usepackage[utf8]{inputenc}
\usepackage{amsmath, amssymb, latexsym}
\usepackage{sidecap}
 
\usepackage{tikz}
\graphicspath{ {Figures}}
\AppendGraphicsExtensions{.gif}

\ifCLASSINFOpdf
\else
\fi
\begin{document}
%
\title{RadDQN: a Deep Q Learning-based Architecture for Finding Time-efficient Minimum Radiation Exposure Pathway}
%
%
%
\author{Biswajit Sadhu,
        Trijit Sadhu,
        and S. Anand}
\thanks{Biswajit Sadhu is with the Health Physics Division, Health Safety and Environment Group, Bhabha Atomic Research Center, Mumbai – 400085, India. He is working as Assistant Professor in the Homi Bhabha National Institute, Mumbai - 400094, India. e-mail: bsadhu@barc.gov.in, biswajit.chem001@gmail.com.}
\thanks{Trijit Sadhu is with Birla Institute of Technology, PILANI. He is pursuing M.Tech in Data Science and Engineering.}
\thanks{S. Anand is with the Health Physics Division, Health Safety and Environment Group, Bhabha Atomic Research Center, Mumbai – 400085, India. He is working as Associate Professor in the Homi Bhabha National Institute, Mumbai - 400094, India.}

\maketitle
  
\begin{abstract}
Recent advancements in deep reinforcement learning (DRL) techniques have sparked its multifaceted applications in the automation sector. Managing complex decision-making problems with DRL encourages its use in the nuclear industry for tasks such as optimizing radiation exposure to the personnel during normal operating conditions and potential accidental scenarios. However, the lack of efficient reward function and effective exploration strategy thwarted its implementation in the development of radiation-aware autonomous unmanned aerial vehicle (UAV) for achieving maximum radiation protection. Here, in this article, we address these intriguing issues and introduce a deep Q-learning based architecture (RadDQN) that operates on a radiation-aware reward function to provide time-efficient minimum radiation-exposure pathway in a radiation zone. We propose a set of unique exploration strategies that fine-tune the extent of exploration and exploitation based on the state-wise variation in radiation exposure during training. Further, we benchmark the predicted path with grid-based deterministic method. We demonstrate that the formulated reward function in conjugation with adequate exploration strategy is effective in handling several scenarios with drastically different radiation field distributions. When compared to vanilla DQN, our model achieves a superior convergence rate and higher training stability.
\end{abstract}


%
\IEEEpeerreviewmaketitle

\section{Introduction}
%
%
%
%
\IEEEPARstart{D}{eep} reinforcement learning (DRL), a sub-branch within Data-Science has become increasingly popular for learning complex behavior in dynamic unstructured environments.\cite{mnih2013playing, mnih2015human, zhu2021bio, li2016traffic, d2024autonomous, chu2022path} Within DRL, an agent interacts with the environment and collects reward or penalties based on its actions. With experience-based knowledge, agent finds optimal policy for attaining desirable behaviour (by maximizing cumulative reward) in the given environment. This ideology behind learning complex environment is well-suited for path-optimization and complex navigation problems involving robots and drones.\cite{jiang2019path} In this context, deep Q networks (DQN), a model-free DRL algorithm,\cite{sutton1988learning} has been successfully applied for various path planning problems involving mobile robots\cite{zhou2018deep, jiang2019path}, unmanned surface \cite{xiaofei2022global} or aerial vehicles,\cite{kong2023b}  coastal\cite{guo2021path} or container ships\cite{gao2023optimized}.  Unlike the popular grid- or graph-based methods, such as Voronoi diagram\cite{bhattacharya2007voronoi}, Dijkstra\cite{dijkstra1959note} or A* algorithm\cite{persson2014sampling}, DQN computationally scales well in high-dimensional decision space.\cite{kulathunga2022reinforcement} Further, by virtue of training, DQN can provide an ensemble of alternate good trajectories allowing user to choose logistically feasible paths based on requirement. 

Path optimization is also central to efficient radiation protection plans.\cite{liu2014path, chao2019dl, zou2018optimization} In case of radiological/nuclear emergency,\cite{onda2020radionuclides} inadvertent release of significant quantities of radionuclide into the environment may cause contamination in large-area. In such scenario, evacuation of people to a safer place may become essential to alleviate the harmful consequence of radiation.\cite{rosoff2007risk, hasegawa2016emergency} In addition to accidental scenario, in routine operations as well\cite{abd2022coverage}, avoidance of radiation hot-spots (indicating high radiation intensity) within a working area is necessary to achieve proper implementation of radiation protection principles like \textbf{ALARA (As Low as Reasonably Achievable)}.\cite{chizhov2017methods} In highly contaminated zone, use of drones or robots\cite{bird2018robot} for radiation mapping can also an reasonable alternative for effective decision making. Undoubtedly, handling these situations require knowledge about the paths that not only has minimum radiation intensity but also time-efficient. Therefore, finding the minimum-radiation pathway that connects the entrance and exit point in a radiation-exposed area is of paramount importance for controlling the radiation dose within prescribed limit\cite{protection2007icrp, liu2015minimum} in order to mitigate the severe deterministic or stochastic effects.\cite{choudhary2018deterministic} However, lack of scientific tools often impede efficient dose-planning and estimation in complex radioactive environment. 

In this context, meta-heuristic bio-inspired particle-swarm optimization (PSO) algorithm was earlier applied to find optimal walking route with distance and radiation dose as decision factors.\cite{liu2014path, wang2018path} Though the obtained path using PSO may leads to minimum radiation exposure but does not guarantee to be time-efficient. Further, issues with the low convergence rate, premature convergence and difficulties on finding global minima in large-dimensional space may restrict application of this method.\cite{persson2014sampling, sucan2012open} Among other approaches, Dijkstra\cite{dijkstra1959note, chizhov2017methods} and A* algorithm\cite{persson2014sampling} was applied in several studies.\cite{alzalloum2010application} Due to the deterministic nature of the algorithm, it produces one solution for the optimum path, which may not necessarily be logistically feasible path owing to the presence of obstacles. Moreover, the exponential scaling in computational cost with large number of grid points also limits the application of these methods in real scenario. Recently, Chao et al. applied  random tree star algorithm in combination with grid-based searching strategy (GB-RRT*) that searches the grid-nodes with obstacles to generate realistic and more practical path.\cite{chao2018grid}

On contrary to the deterministic methods discussed above, DQN allows training of agent in static or dynamic environments to help it learn the optimal strategy for stochastic decision-making, thereby facilitating its application in autonomous decision-making field.\cite{xiaofei2022global} Further, DQN-trained agent can learn obstacle-avoidance property to put forward alternate trajectories.\cite{lei2018dynamic} However, to the best of our knowledge, the application of DQN or DRL for mapping the minimum radiation-exposure path is not reported so far. The major bottleneck could be the lack of radiation-aware reward function that enables the agent to learn the complex radioactive environment. Further, unlike the obstacle-avoidance problem, avoiding radiation exposure during path-planning is tricky as the involve task not only requires staying away from the radiation source but also aim for paths that are maximally distant from the radioactive source to minimize the cumulative radiation-exposure. Undoubtedly, this becomes even more complicated when multiple radiation sources are present within the contaminated zone. Moreover, as we discuss later, the stability of training, issues with convergence and trade-off between exploration and exploitation in simulated environment often take heavy toll on practicability of DQN.\cite{fan2020theoretical, anschel2017averaged}    

On this background, here we introduce a DQN-based algorithmic framework named \textbf{RadDQN} that addresses the aforementioned issues and designed to provide an
optimum time-efficient escape-route with minimum radiation exposure in a simulated zone contaminated by radioactive materials. We simulate the contaminated zone using 2D Grid-World, representing the floor area of the zone with an entry and exit cell, where \textit{n} numbers of radioactive sources placed randomly at arbitrary locations on the floor surface. We formulate an efficient reward structure which is aware of the presence of multiple radiation sources and their radiation strength of the sources, the duration of exposure and radial distance of the agent from all the radiation hot-spots and the destination point. Further, we propose unique strategies on controlling the exploration/exploitation based on simulation environment to achieve superior time-efficient convergence with reduced variance than the traditional DQN with $\epsilon$-greedy method. The structure of this paper is organized as follows. Section II introduces the framework of our approach, training protocol, exploration/exploitation strategies; Section III discusses the finding of simulation experiments with varying number of radiation sources and their strengths; and finally, Section IV provides the important inferences about this work and suggests probable future works.

\section{Methodology}

The methodology used in this work for training the agent considerably differs from the traditional DQN. Here in this section, we first briefly describe the existing methodologies and working principles of DQN. Later we describe the prescribed changes to obtain RadDQN framework. Further, we devote a subsection to discuss the formulation and algorithm used for finding optimal path and methodology for comparison with RadDQN generated ensemble of paths.    

\subsection{Reinforcement Learning and Vanilla DQN}

Reinforcement learning (RL) algorithm\cite{sutton1988learning} deals with the agent and the environment. A task is given to the agent, which it performs by interacting with the environment through different set of actions. With every action, agent receives incentives/rewards. In this regard, a $Q$-table helps agent choosing best possible action, and correct the values in $Q$-table upon trial-and-error based on the reward it has received. RL follows the formal framework of Markov Decision Process (MDP). Therefore, the agent’s future only depend on the current state.\cite{jia2020review} As a result agent maps its individual actions to learn an optimal function leading to maximum incentives or rewards. Over the course of several attempts (or episodes), agent optimizes the state-based best action and figures out the optimal strategy that maximizes the cumulative reward.\cite{sutton1988learning} 


For instance, assume that an agent executes an action \textit{a} in a state \textit{s} and get a reward \textit{r} in return from the environment. The $Q(s,a)$ value denotes the quality of that action. Every such action results in new and updated $Q(s,a)$ values based on the state of the agent (Figure \ref{fig:flowchart_dqn}). Subsequently, updation of the $Q$-table takes places according to the \textit{Bellman equation}:

\begin{equation}
\tag{1}
Q(s, a) \xleftarrow{} Q(s, a) + \alpha[r + \gamma \max\limits_{a'}Q(s', a') - Q(s, a)]
\end{equation}

where \textit{$a^\prime$} denotes the chosen action in the next state \textit{$s^\prime$}. $\alpha$ is the learning rate. $\gamma$ is the discount factor which controls the importance of present action and reward in the context of achieving future. $\max\limits_{a'}Q(s', a')$ is the maximum cumulative reward value corresponding to state $s^\prime$.

In DQN, the $Q$-table is replaced with the function-approximator or neural network to facilitate the decision making in complex scenario involving significant number of states and actions.\cite{mnih2013playing, mnih2015human} As DQN solely relies on the experiences of the agent, it is model-free off-policy reinforcement algorithm in nature. The agent's experiences in the form of [$s$, $a$, $r$, $s^\prime$] obtained (during the training session) work as data samples for the future run. However, DQN suffers from two major issues. Firstly, training-led experiences may become highly correlated to question the validity of MDP. Secondly, issues with the convergence of value function. Both these problems are addressed by Google's Deep-mind in their pioneering work on playing atari games using DQN.\cite{mnih2013playing, mnih2015human} The experience replay, which allows random mini-batch sampling from the pool of experiences namely \textit{replay memory} $D$, removes the correlation among the samples while the use of \textit{target network} provides stability on the convergence by giving user-specific control over the iterative updation of value function. This further helps in dealing with \textit{``catastrophic forgetting''}.\cite{mnih2013playing} The flowchart of DQN is given in supporting information (Figure \ref{fig:flowchart_dqn}).    

Essentially, the agent's choice on the current action can leverage the existing knowledge in terms of past experiences (i.e. \textit{exploitation}). Or, agent can take an action that is never tried in search of finding better result (i.e. \textit{exploration}). In this context, balancing between these two possibilities are required to efficiently explore larger state-space for finding the optimal policy quickly. In this regard, local dithering techniques like $\epsilon$-greedy algorithm is commonly used to allow large exploration (with probability \textit{$\epsilon$}) at the beginning of the training phase and later increases exploitation with increasing number of training episodes. In recent studies,\cite{george2022deep, pu2022deep, fotouhi2021deep} DQN is often used with the aforementioned tricks and trades (experienced replay, target network with $\epsilon$-greedy algorithm). Collectively, we refer to them as \textbf{vanilla DQN} for the rest of the article. 

\subsection{Our Approach: RadDQN}

\begin{figure*}[htbp]
    \centering
    \includegraphics[width=\textwidth]{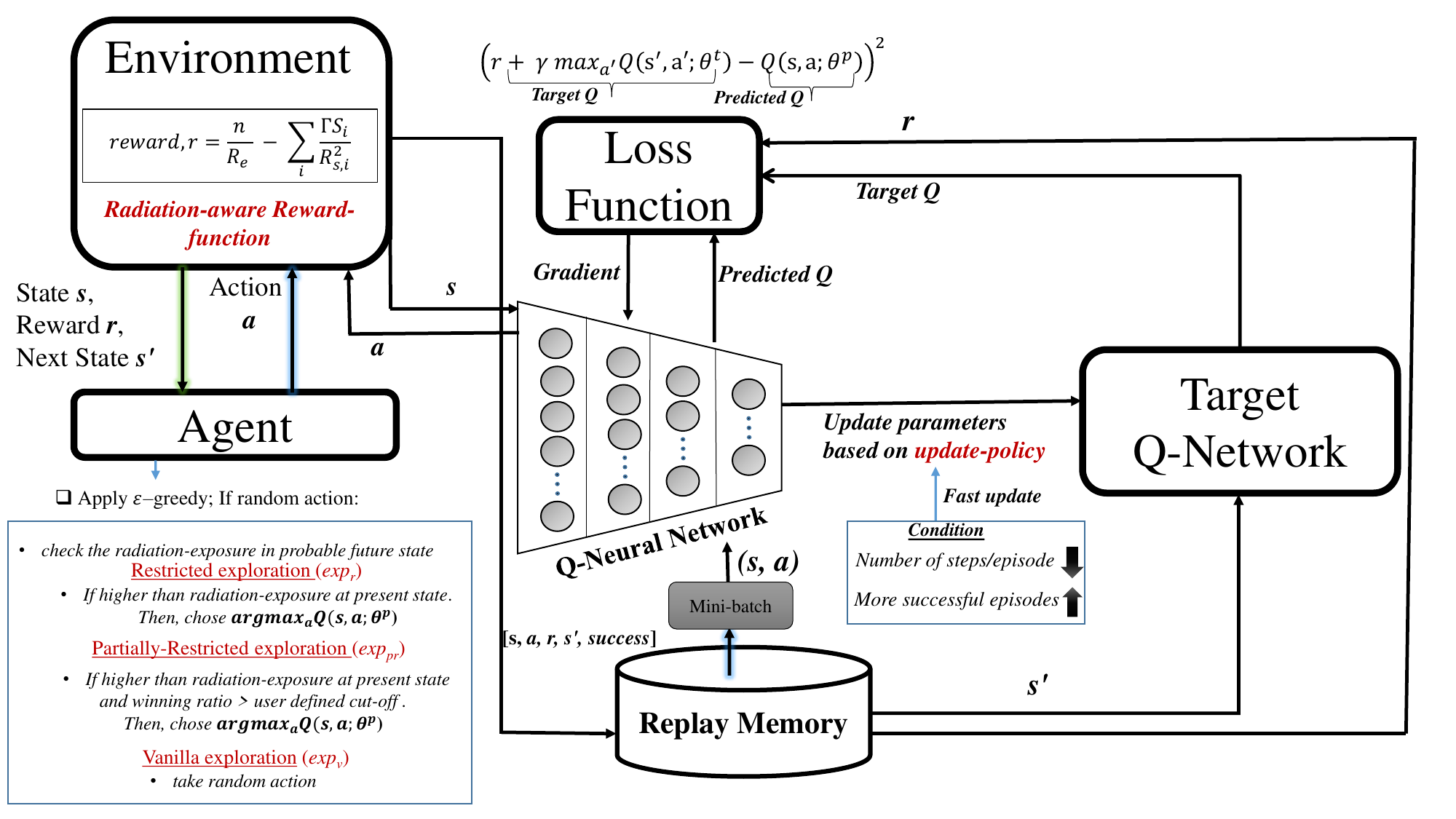}
    \caption{Flowchart of RadDQN algorithm (present study). }
    \label{fig:flowchart_raddqn}
\end{figure*}

Keeping the objective of training in mind, here we first describe the simulated environment and then delineate the modifications in \textbf{vanilla DQN} to obtain \textbf{RadDQN} (Figure \ref{fig:flowchart_raddqn}). In robotics and path-planning operations, often the agents are trained with the task of reaching destination in minimum time while avoiding the space in which some obstacles resides. Note that radiation exposure from radioactive materials are not limited to some defined space as it can radiate spaces far away from the actual source location depending on its radiation strength/intensity. The presence of multiple radiation sources further complicates the distribution of radiation. Considering these issues, we perform following changes in the vanilla DQN architecture for usage of DQN in the radiation-based path-planning problem. In the later section, we will further show how such changes help to overcome the challenges during training.

\subsubsection{Simulation Environment}

We simulate the contaminated zone using two-dimensional grid world with an entry and exit cell. This represents the floor area of the zone where \textit{n} numbers of radioactive sources may be placed randomly at arbitrary locations on the floor surface. While 10 $\times$ 10 square-grid (i.e. 100 states) is chosen for the purpose of demonstration, much larger space can be considered in the developed algorithm. In order to move toward the exit cell, the agent in a cell can possibly choose any action from 8 valid actions : up, down, left, right and four diagonal moves (upper right, upper left, down right, down left). If the agent hits the boundary of the wall during training, the action is considered invalid. In such case, the agent is asked to choose the next available valid action. As revisiting the visited state is undesirable considering the cumulative increase in radiation exposure, agent is not allowed to revisit any previously visited state. Therefore, the list of valid actions in a cell must lead to states that are not visited during that episode. Considering the available number of states and applied scenarios, agent is allowed to take maximum of 30 steps to reach the destination. If the number of steps is more than 30 or the agent reached a state with no valid actions, the episode is ended with failure label. To investigate the training with diversified scenarios, we placed 2 or 3 radioactive-sources of same or variable strength in meaningful locations. Further details on the strength and spatial position positions are discussed in the result (section III).       

\subsubsection{Radiation-aware Reward Function}

\begin{figure}[htbp!]
    \centering
    \includegraphics[width=\linewidth]{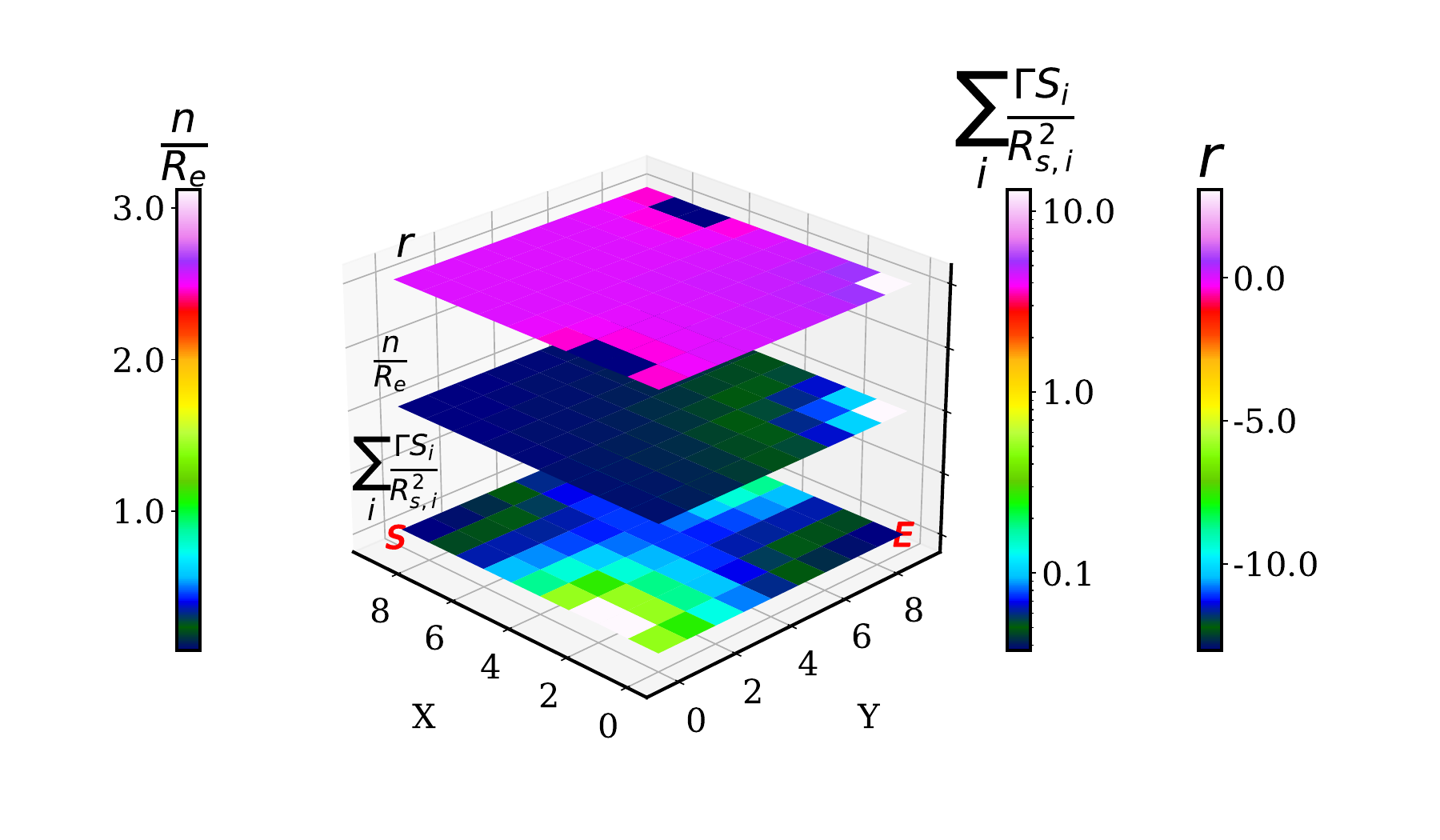}
    \caption{Radiation-aware reward function for two sources of unit radiation strength at (2,0) and (7,7). Reward \textbf{r} results from the subtraction of $\frac{n}{R_{e}}$ from $\sum_i{\frac{\Gamma S_i}{R_{s,i}^{2}}}$ following equation \ref{eq:reward}. In the figure, the start and exit cell are symbolized as `S' and `E', respectively. The value of $\Gamma$ and n are taken as 1.}
    \label{fig:reward_function}
\end{figure}

The reward-function essentially drives the learning process of the agent. Therefore, formulating better reward-functions has been an active topic in path-planing problem\cite{guo2021path,du2021optimized} and central to any DRL architecture\cite{icarte2022reward}. The agent screens the action based on the reward structure, which makes these reward function more specific to the objective at hand. In the present problem, to make the agent aware of the position and intensity of radiation-flux, we crafted the reward structure following the principle of radiation exposure.\cite{kim2018three} This is achieved by applying inverse-square law (\textbf{$\frac{\Gamma S}{R_{s}^{2}}$}), which states that the radiation intensity is inversely proportional to the square of the radial distance from all individual sources (\textbf{R$_{s, i}$}). Here, \textit{i} corresponds to the number of sources present in the given environment and $\Gamma$ denotes the specific Gamma ray constant.\cite{peplow2020specific} In the present study, the value of $\Gamma$ is considered as \textbf{1}. The effect of radiation strength of the source is accounted by keeping source-strength \textbf{S} in the numerator. However, the mere consideration of radiation exposure in the state-space is not good enough for the task of finding minimum exposure pathway as it does not contain knowledge about the proximity of the agent from the destination (\textbf{R$_e$}). Addressing this is essential because in the absence of it the agent might prefer to stay at the region of low radiation flux for longer time rather than urging to reach the destination. To alleviate this problem, we assigned a positive reinforcement equivalent to \(\frac{n}{R_{e}}\) to the agent.(cf. Equation \ref{eq:reward} and Figure \ref{fig:reward_function}). As the radiation-exposure is a positive quantity in the state-space, this makes the overall state-based reward (\textbf{r}) in  equation \ref{eq:reward} mostly negative unless positive reward gained through the closeness to the exit outweighs the exposure. This is particularly true for environments with very high radiation field. Figure \ref{fig:reward_function} depicts the contributions of each of the attributes into the resultant radiation-aware reward function. Note that, the parameter \textit{n} in the equation has the controlling role on prioritizing the objective of reaching destination. With increasingly higher value, agent develops higher urge to reach exit while low values of \textit{n} help agent to focus on the finer dose distribution in the spatial domain. Therefore, \textit{n} may be considered as a key variable that can modulate the shape of the reward function as per the need of operation. For the present study, \textit{n} $=$ 1 is considered.          

\begin{equation*}
\tag{2}
r = \frac{n}{R_{e}} - \sum_i{ \frac{\Gamma S_i}{R_{s,i}^{2}}}
\label{eq:reward}
\end{equation*}

\subsubsection{Defining Optimum Path}

In this work, we define a path optimum if it leads to minimum collective radiation exposure to the agent and requires minimum number of steps to reach the exit. Note that the velocity at which the agent travels is assumed constant, which makes the number of steps equivalent to number of time steps. Considering this, the optimum path determination essentially boils down to the minimization of the rate of exposure (i.e. cumulative reward/number of steps) for the batch of episodes. The training process provides the density of visited states and an ensemble of trajectories. The optimum or near-optimum paths may be obtained by connecting the most visited states of the agent at the end of the training upon convergence.

\subsubsection{Neural Network (NN) Configuration}

As mentioned earlier, the simulated environment consists of one entrance and exit at the opposite corner of the grid (here (0,9) and (9,0) in 2D grid, respectively) (Figure \ref{fig:two_sources}). For simulated environment consisting one agent, \textit{i} number of radiation sources and \textit{t} number of states, the input has dimension of \textit{(i+1)} $\times$ \textit{t}. Our neural network performs affine transformation on the the input dimension with two successive linear layers of dimension 200 and 100, respectively. Later, the non-linearity is introduced with LeakyRelu activation function\cite{maas2013rectifier}. The final output is the \textit{Q} values for eight actions. The optimization is performed using Adam optimizer\cite{kingma2014adam} without any weight decay. The weights of NN is initialized with Xavier normalization.\cite{glorot2010understanding} The training is carried out using PyTorch deep-learning library using mean square error (MSE) as loss function.\cite{NEURIPS2019_9015} Table \ref{tab:t_param} tabulates the important training parameter.

\newcolumntype{Y}{>{\centering\arraybackslash}X}
\begin{table}[]
\caption{Training Parameters of RadDQN}
\centering
\begin{tabularx}{\columnwidth}{@{} Y | Y @{}}
\hline
Training Parameters    & Value \\ \midrule
Learning rate          & 0.001 \\
discount factor        & 0.9   \\
mini-batch size        & 30    \\
replay buffer          & 2000  \\
number of epoch        & 5000  \\
update frequency$^{a}$ & 600   \\ \bottomrule
\end{tabularx}
\label{tab:t_param}
\footnotesize{
$^a$Update frequency is user-specific and varies following algorithm \ref{alg:sync}.}
\end{table}

\subsubsection{Exploration vs. Exploitation}

Developing strategies for efficient random exploration has been an active topic of research. Despite significant advancement on different exploration options such as Boltzmann exploration and entropy regularization\cite{peters2010relative}, dithering techniques like $\epsilon$-greedy is generally preferred due to its simplicity and easy-adaptation on complex problems. $\epsilon$-greedy strategy treats all the actions in equal footing making the learning slow in large area exploration. Proposed methods such as temporally-extended $\epsilon$-greedy exploration considered repetition of same actions for fixed duration to improve the outcome of $\epsilon$-greedy exploration in the context of persistence.\cite{dabney2020temporally} Jiang et al. restricted the blind exploration through incorporation of goal-directed heuristic knowledge into DQN architecture.\cite{jiang2019path} Here, we pay our attention on controlling the exploration by introducing the \textit{condition-based conversion of the random action into model-directed action}. We demonstrate later that the stabilization of training is closely tied to the agent's exploration and exploitation abilities. The following three strategies are used to test the agent's performance and path convergence. (cf. Algorithm \ref{alg:exploit}).

(a) Vanilla exploration strategy (\textit{exp$_{v}$}): Here, we relied on the $\epsilon$-greedy algorithm that allows exploration with probability $\epsilon$ at the beginning of the training while the extent of exploitation increases with increasing number of episodes.
 
(b) Restricted exploration strategy (\textit{exp$_{r}$}): This strategy restricts blind exploration by allowing the agent to sneak-peak at the reward in a probable future state. This is applied only in those situations where $\epsilon$-greedy algorithm suggested to take random actions. More precisely, instead of following the $epsilon$-greedy algorithm's random action, the agent is instructed to follow model-directed (NN) action if attaining a particular future state leads to higher radiation-exposure than its current state. Therefore, \textit{exp$_{r}$} converts the random action into model-directed one based on characteristics of future reward. This is supposed to provide NN more opportunities to refine action values based on past experiences.
 
(c) Partially-restricted exploration strategy (\textit{exp$_{pr}$}): Here, we use the same protocol as the \textit{exp$_r$} strategy to convert random actions to model-directed ones, but with more specific and stricter condition. It considers the radiation-exposure at future state as well as the winning ratio of played episodes. Therefore, \textit{exp$_{pr}$} partially restricts the blind exploration allows relatively more number of random actions in training. However, we do apply a cutoff threshold on winning ratio (of played episodes) for activating this exploration strategy below which agent continues to perform random exploration following $epsilon$-greedy algorithm. Section III provides more finer details on the distribution of random and model-directed actions upon change of exploration strategy.       

\RestyleAlgo{ruled}

\SetKwComment{Comment}{/* }{ */}

\begin{algorithm}
\caption{Modified strategy on Exploitation vs. Exploration}\label{alg:exploit}
\KwData{ $Q$ = action-value function

$s$ = current state;

$N_{ep}^{min}$ = minimum number of episodes to consider update;

$N_{win}^{min}$ = minimum number of winning episodes;

$R_{win}^{average}$ = average ratio of win in last k episodes during training ($k$= 10 in current study);


}
\KwResult{Decide whether to explore or exploit}

\For{$episode(i)$}{

\For{$step$}{

$n \gets $ uniform random number between 0 and 1;

{\If{$n$ $<$ $\epsilon$} 
{\If{$exp_v$}{
choose random action $a$ from action space.\ \Comment*[r]{vanilla exploration}} 
\Comment*[r]{partially restricted exploration}
{\ElseIf{$exp_{pr}$} 
{
\eIf{$i$ $>$ $N_{ep}^{min}$ and $R_{win}^{average}$ $>$ $R_{win}^{min}$ and $r_{current step}^{\max\limits_{a}Q(s, a)}$ $>$ $r_{previous}$}{action $a \gets $ $\max\limits_{a}Q(s, a)$} 
{choose random action $a$ from action space.}}}
\Comment*[r]{restricted exploration}
{\ElseIf{$exp_r$} 
{
\eIf{$r_{current step}^{\max\limits_{a}Q(s, a)}$ $>$ $r_{previous}$}{action $a \gets $ $\max\limits_{a}Q(s, a)$} 
{choose random action $a$ from action space.}}}}
}
}
}
\end{algorithm}

\subsubsection{Update frequency of Target Network}

The updation of DQN weights in every time steps can destabilize the convergence. Target network restricts such frequent updation. It essentially keeps a copy of main DQN network that is delayed by some user-defined number of steps (i.e. update frequency). This allows incremental changes in the DQN parameters thereby helps in the convergence. Update frequency is a variable that can be only optimized based on many trial-and-errors. The standard practice is to update the target network periodically, therefore making the update frequency constant over the course of training. The theoretical explanation on how and why target network stabilize the training is still an open topic of investigation. Recent work by Fellows et al.\cite{fellows2023target} showed that the proper tuning of update frequency may help achieving convergence and training stability. Inspired by this, we make this variable dependent on two factors, namely winning ratio of all played episodes and change in the moving average of number of steps during last k winning episodes as compared to all played episodes (cf. Algorithm \ref{alg:sync}). If the agent's performance improves over the course of training, algorithm \ref{alg:sync} allows faster updation of target network so that the accuracy gained by the parent network in last few episodes is transferred to target network leading to faster learning. Whereas,  algorithm \ref{alg:sync} ensures slower update (controlled by update factor \textit{uf}) if the agent's performance worsen or doesn't improve over the time.


\RestyleAlgo{ruled}

\SetKwComment{Comment}{/* }{ */}

\begin{algorithm}
\caption{Improvising the update frequency of target network}\label{alg:sync}
\KwData{$N_{ep}^{min}$ = minimum number of episodes to consider update;

$S_{f}$ = user-defined update frequency

$N_{ep}^{min}$ =  Minimum number of episodes to wait before the algorithm starts.

$N_{win}$ = number of winning episodes;

$N_{win}^{min}$ = minimum number of winning episodes to consider an update;

$uf$ = update factor for slow update ($>$ 1);

$N_{steps}^{win}$ = moving average of number of steps of all trained episodes excluding last m episodes;

$N_{steps}^{last-k-episodes}$ = moving average of number of steps in last k winning episodes;

$R_{win}^{current}$ = ratio of win till the last episode during training.

$R_{win}^{previous}$ = ratio of win till the penultimate episode during training.
}
\KwResult{obtain update frequency based on performance of agent}

$N_{steps}$ = 0

\For{$episode(i)$}{

\For{$step$}{
$N_{steps}$ = $N_{steps}$ + 1

\If{number of episodes $>$ $N_{ep}^{min}$ }{
$S_{f}$ = $S_{f}$ (1 - $R_{win}^{current}$)\ 

\If{$N_{steps}$ $>$ $S_{f}$} {\If{$N_{win}$ $<$ $N_{win}^{min}$}
{update the target network.\ \Comment*[r]{fast update} 

$N_{steps}$ = 0
}
\ElseIf{$N_{win}$ $>$ $N_{win}^{min}$ and $R_{win}^{current}$ $>$ $R_{win}^{previous}$ and $N_{steps}^{win}$ $>$ $N_{steps}^{last-k-episodes}$}{

update the target network\ \Comment*[r]{fast update} 

$N_{steps}$ = 0
  }
  
\ElseIf{$N_{steps}$ $>$ $uf$ $\times$ $S_{f}$} {update the target network\ \Comment*[r]{slow update} 

$N_{steps}$ = 0}

}{no update}} 
}
}
\end{algorithm}

\subsection{Finding ground-truth:}

Analysing the performance of RadDQN warrants information on ground-truth of optimum path. Analytical approaches like solving the Euler-Lagrange equation on radiation-aware reward function (eq. \ref{eq:reward}) may provide the true path. However, particularly for environments with multiple sources, finding such solution is often computationally intractable\cite{shen2005optimal} due to the complex nature of the objective function.\cite{vernaza2010learning} Instead, here we compute the true optimized path using graph-theory based Dijkstra algorithm that finds the shortest path between any two nodes/vertices in a edge-weighted graph.\cite{dijkstra1959note} To apply it, we convert the spatial environment (2D grid) into a graph where each grid points can be considered as vertex. The edge is the path connecting any two adjacent vertices. Later, we compute the radiation exposure at every vertex using inverse-square law ($\sum_i{\frac{S_i}{R_{s,i}^{2}}}$) for $i$ number of sources at present at $R_{s,i}$ distances. The radiation exposure at the edges are calculated by averaging the radiation-intensity values of connected vertices. Finally, Dijkstra's algorithm is applied on the graph using python-based networkx package.\cite{hagberg2020networkx}   

\subsection{Comparison of ground-truth with RadDQN-predicted paths:}

The training using RadDQN architecture generates an ensemble of paths. To study how these generated paths compare with the ground truth, we compute Fr\'echet distance,\cite{brakatsoulas2005map} which measure the similarity of NN generated paths with true-path based on the location and ordering of the grid points along the path. Consider any two trajectories (say, P and Q) with m and n number of grid points, respectively. Here, Fr\'echet distance ($\delta_F$) is defined by the minimum cord-length that allows traversing both of these trajectories from beginning to end (equation \ref{eq:frechet}).  

\begin{equation*}
\tag{3}
\delta_F (P,Q) = \inf_{\alpha, \beta} \max_{t \in [0,1]} || P(\alpha(t)- Q(\beta(t)||
\label{eq:frechet}
\end{equation*}

$\alpha$ and $\beta$ refers to all possible reparametrizations with range [0,1]. Further details on this metric is available elsewhere.\cite{brakatsoulas2005map, driemel2022discrete} 

\section{Results and discussion}

Training a robot in real scenario involving thousands of trial and error is a daunting task. This is especially difficult with the radiation sources in place considering the safety of the personnel in the training and for dealing issues with decontamination of robot systems. Therefore, here we rely on the simulated environment that mimics the reality as much as possible. This is to achieve a theoretical foundation with good accuracy on path-planning algorithms in simulated environment before it can be tested in the actual experiment. The details of the simulated environment are presented in the preceding section. Here, 2D plane can be considered mimicking the floor of a room. It has one entrance (\textbf{\textit{S}}, (0,9) position) and one exit point (\textbf{\textit{E}}, (9,0) position) (Figure \ref{fig:two_sources}). The agent always start its journey from the entrance with the aim of reaching exit point avoiding the radiation exposure from the radioactive materials kept on the floor. In the process, it acquires reward/penalty based on radiation-aware reward function (Eq. \ref{eq:reward}). Here, at first we discuss the cases where two radioactive sources are kept on the floor. In the later part of this section, we make the scenarios more challenging with more number of sources of variable radiation strength. Furthermore, we compare and analyze the performance of RadDQN against the vanilla DQN approach in the context of applied modifications.       

\subsection{Scenario with two radioactive sources}

\begin{figure*}[htbp!]
    \centering
    \includegraphics[scale=0.7]{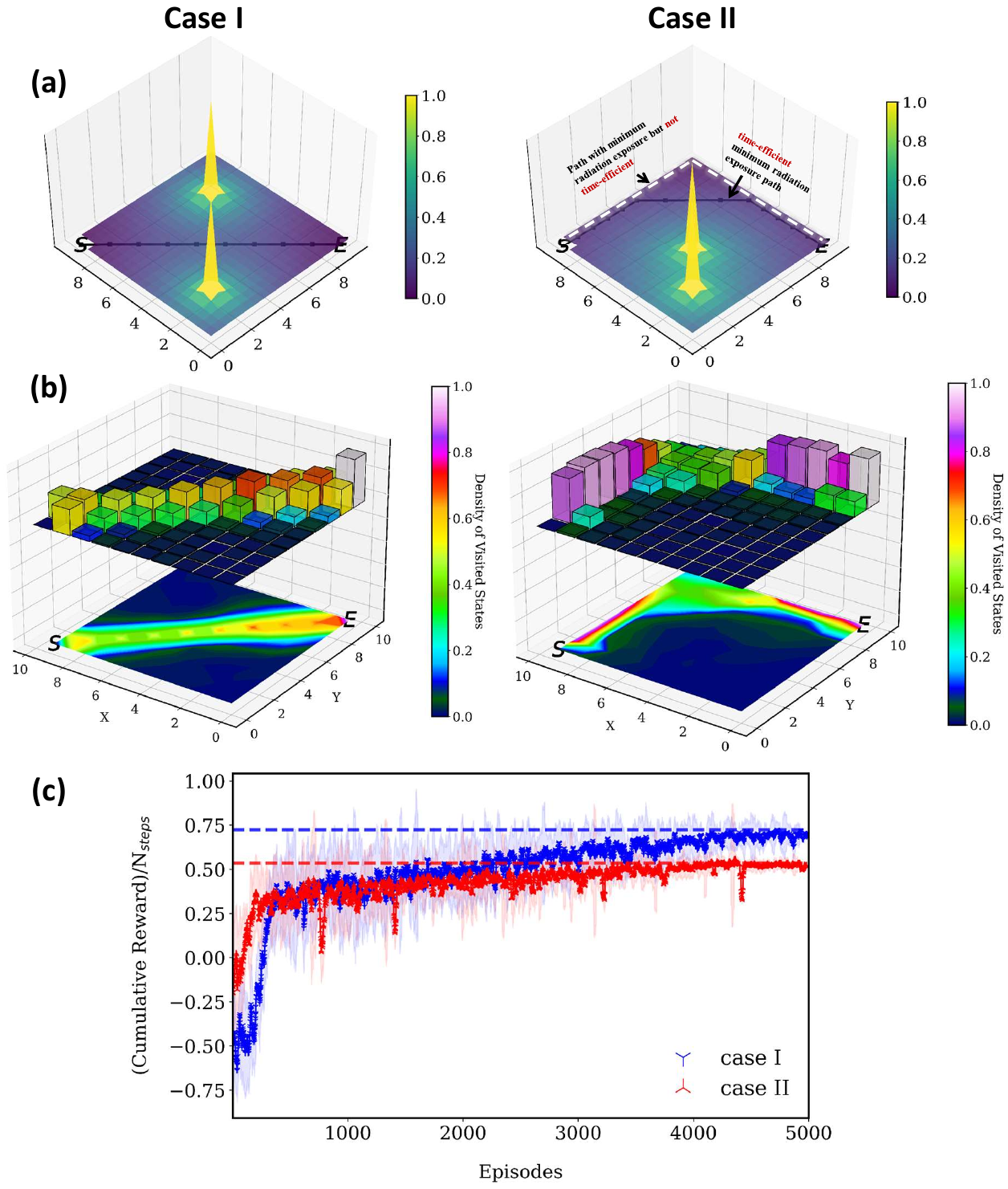}
    \caption{The result for the two different environments (Case I and Case II) with two sources. The start and exit point in the simulated floor is shown using `S' and `E' symbol. Top panel (a) shows the distribution of radiation exposure in the simulated floor. The colorbar indicates the intensity of the radiation. The white dashed line in right figure (Case II) indicates the path connecting the states with least radiation exposure (but with more number of steps). The optimal path (ground truth) obtained from Dijkstra's algorithm is shown in black line, where the black dots in the path indicates the number of steps. Middle Panel (b) shows the density of visited states using RadDQN architecture. The associated contour plot in the figure clearly shows the optimal path similarity with  the ground truth. Bottom panel (c) shows the convergence of training in the context of maximizing average reward per step. The dashed line indicates the ground truth value.}
    \label{fig:two_sources}
\end{figure*}

In \textbf{Case I}, two radioactive sources of equal strength are placed at (2,2) and (7,7) respectively (Figure \ref{fig:two_sources}a). Evidently, the diagonal path connecting the entrance (\textit{S}) and exit (\textit{E}) is the best way to reach the exit with minimal radiation exposure in this scenario. Indeed, agent could find this path efficiently. Note that in case I, the diagonal path is not only the path with minimum radiation exposure but also the shortest path (by radial distance) to reach \textit{E}. Therefore, the predicted path isn't a sufficient proof that the agent has understood the notion of radiation-exposure. In this regard, we consider a different scenario (\textbf{Case II}) where two sources of equal strength are kept at (2,2) and (4,4) position, such that the diagonal path (shortest path in terms of number of steps) have more radiation exposure. In this environment, agent learns to avoid the diagonal path and finds a trajectory on the right side of the grid-plane to reach \textit{E}. Note that the agent may have chosen a path along the boundary wall of grid plane (shown in white dashed line, Figure \ref{fig:two_sources}a) that is farthest from the radiation sources. However, this path would require more steps, making it time-consuming to reach \textit{E}. Importantly, the agent's prediction is to take a slightly diagonal route near the rightmost corner, which will reduce the number of steps and reduce the total exposure. The density of visited states during the training period is shown in Figure \ref{fig:two_sources}b, which nicely followed the optimal path obtained through Dijkstra's algorithm (Figure \ref{fig:two_sources}a). In both cases, the training of NN converged nicely towards the ground truth (Figure \ref{fig:two_sources}c). The progress in terms of winning percentage and number of steps required to exit provides an interesting insight into the agent's learning priorities (supporting information, Figure \ref{fig:two_sources_winmoves}). Early in the training, the agent appears to learn quickly about how to reach the exit making winning as priority. While, at the later stage of training, to achieve the optimal solution, it weighs the contribution of radiation exposure in the learned trajectories.  

\subsection{Scenario with three radioactive sources}
Here we further increase the number of sources in the simulated floor and strategically place them to build three different scenarios. This is to check a) the performance of the agent in an environment where there are multiple optimal paths and b) to examine if the placement of radioactive sources near the entrance and exit is detrimental to the agent's learning. Additionally, we investigate the agent's capacity to learn when exposed to radioactive sources with various radiation strengths. 

\begin{figure}[htbp!]
    \centering
    \includegraphics[width=\linewidth]{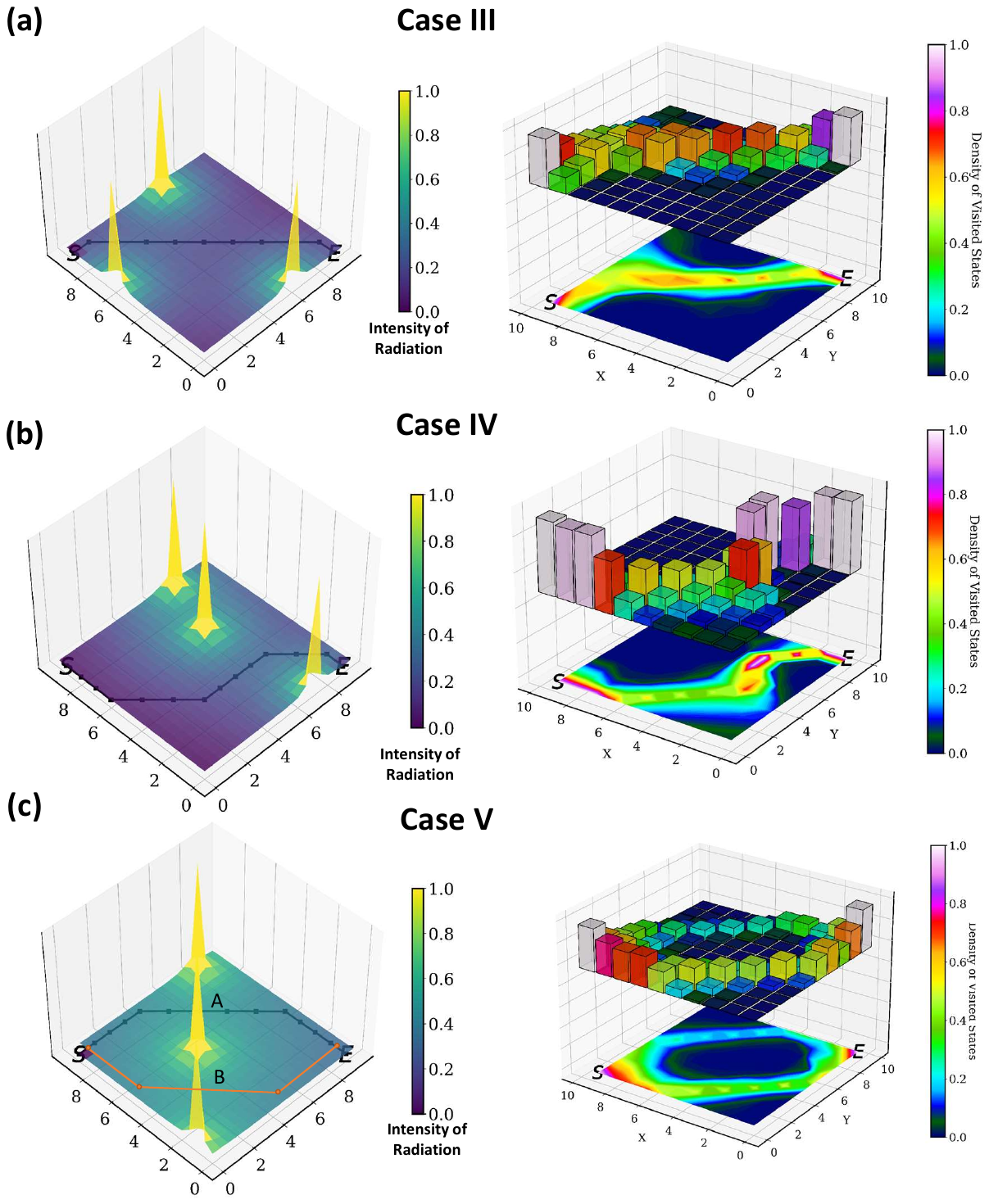}
    \caption{The case-studies with three sources at strategic positions. The start and exit point in the simulated floor is shown using `S' and `E' symbol. Left panel (a, b and c) shows the distribution of radiation exposure in the simulated floor. The colorbar indicates the intensity of the radiation. The optimal path (ground truth) obtained from Dijkstra's algorithm is shown in black line, where the black dots in the path indicates the number of steps. Right Panel (a, b, c) shows the density of visited states using RadDQN architecture. The associated contour plot in the figure clearly shows the optimal path similarity with the ground truth.}
    \label{fig:three_sources}
\end{figure}

\subsubsection{Varying position of radioactive sources}

\textbf{Case III:} Here, one of the sources of relatively higher strength are kept close to \textit{S} (Figure \ref{fig:three_sources}a). At the start of the episode, the agent experiences a high negative reward or penalty due to the presence of a highly radioactive source. The test was designed to evaluate the agent's ability to locate the minimum radiation exposure zone far away from the entrance. 

\textbf{Case IV:} Unlike case III, we looked at a scenario where the area near \textit{E} experiences high radiation exposure, and the area near \textit{S} experiences significantly low radiation. The objective of case IV was to examine whether the agent opts to avoid exiting to remain in the low-radiation zone around the entrance. Furthermore, it is apparent that the placement of radioactive sources necessitates the agent to take crucial diversions to locate the true path proposed by Dijkstra's algorithm. Here also, we observed quick convergence of NN under the \textit{exp$_{pr}$} setting. We analyze the performance of other settings later in this section. 

\begin{figure}[htbp!]
    \centering
    \includegraphics[width=\linewidth]{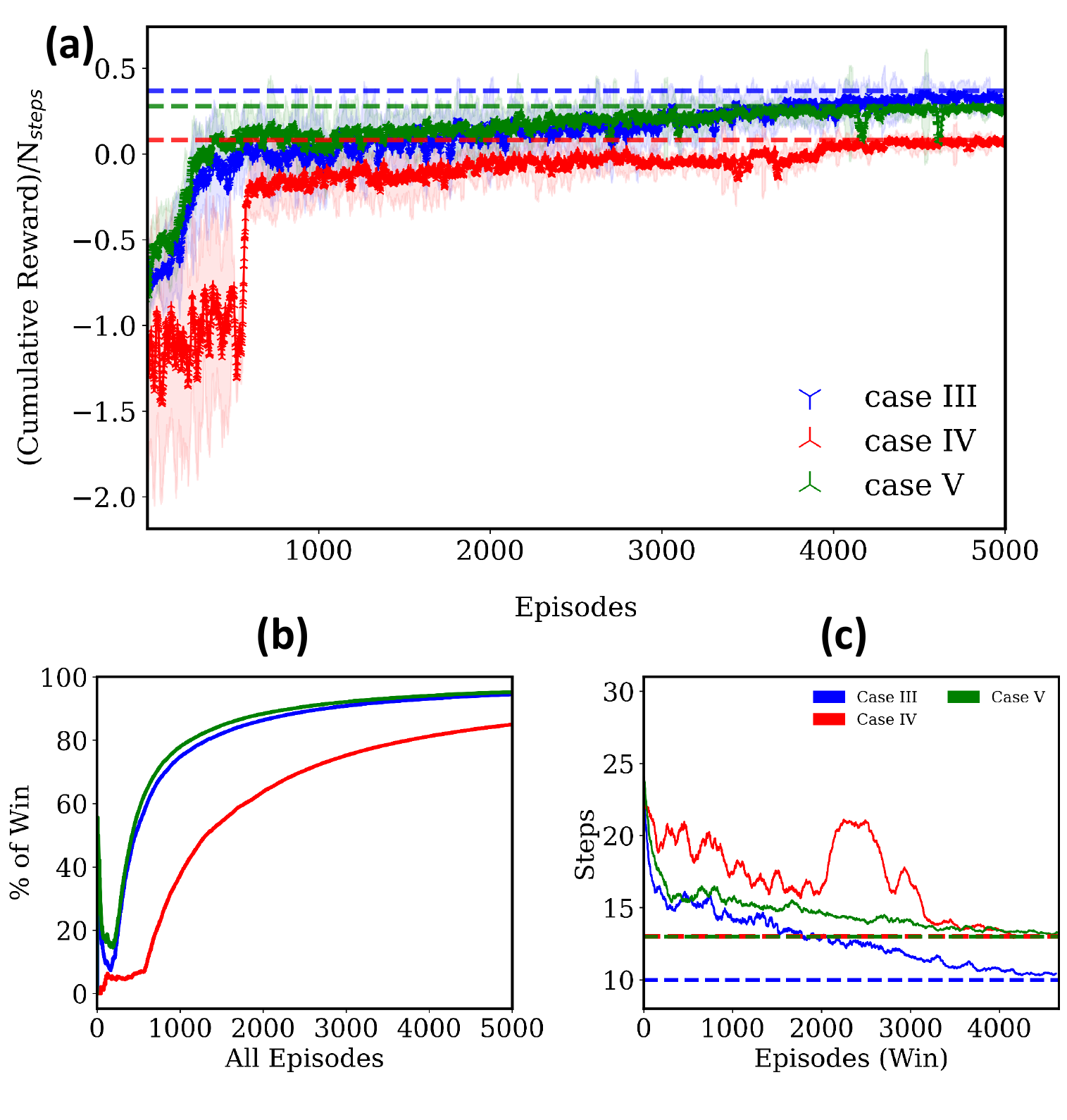}
    \caption{(a) Plot of average cumulative reward vs. number of played episode that shows the convergence of training in the context of maximizing average reward per step. The dashed line indicates the ground truth value (Top panel). (b) Plot of \% of win vs. Episodes (bottom left panel). (c) Plot of number of steps in the winning episodes vs. Index of Winning Episodes (bottom right panel).}
    \label{fig:three_sources_reward_compare}
\end{figure}

\textbf{Case V:} Here, the arrangement of sources is such that two possible solution exists regarding the optimal path. In reality, such situations may commonly arise where multiple path could provide similar cumulative radiation exposure. Finding all possible paths is essential for an agent to make the best decision, taking logistics and other aspects into account. Figure \ref{fig:three_sources}c depicts one such scenario where the paths A or B lead to minimum exposure with equal number of steps, therefore both paths are the optimal solution. In this circumstance, the Dijkstra's algorithm provides a single solution (path A, in Figure \ref{fig:three_sources}c). While, the agent's effective exploration ability is indicated by the fact that a RadDQN-trained agent finds both paths during training.

Evidently, regardless of the radioactive source's location, the training in RadDQN is seen to have good congruence with the ground truth both in the context of cumulative reward and number of steps (Figure \ref{fig:three_sources_reward_compare}).

\subsubsection{Varying radiation strength of sources} 

\begin{figure*}[htbp!]
    \centering
    \includegraphics[width=\textwidth]{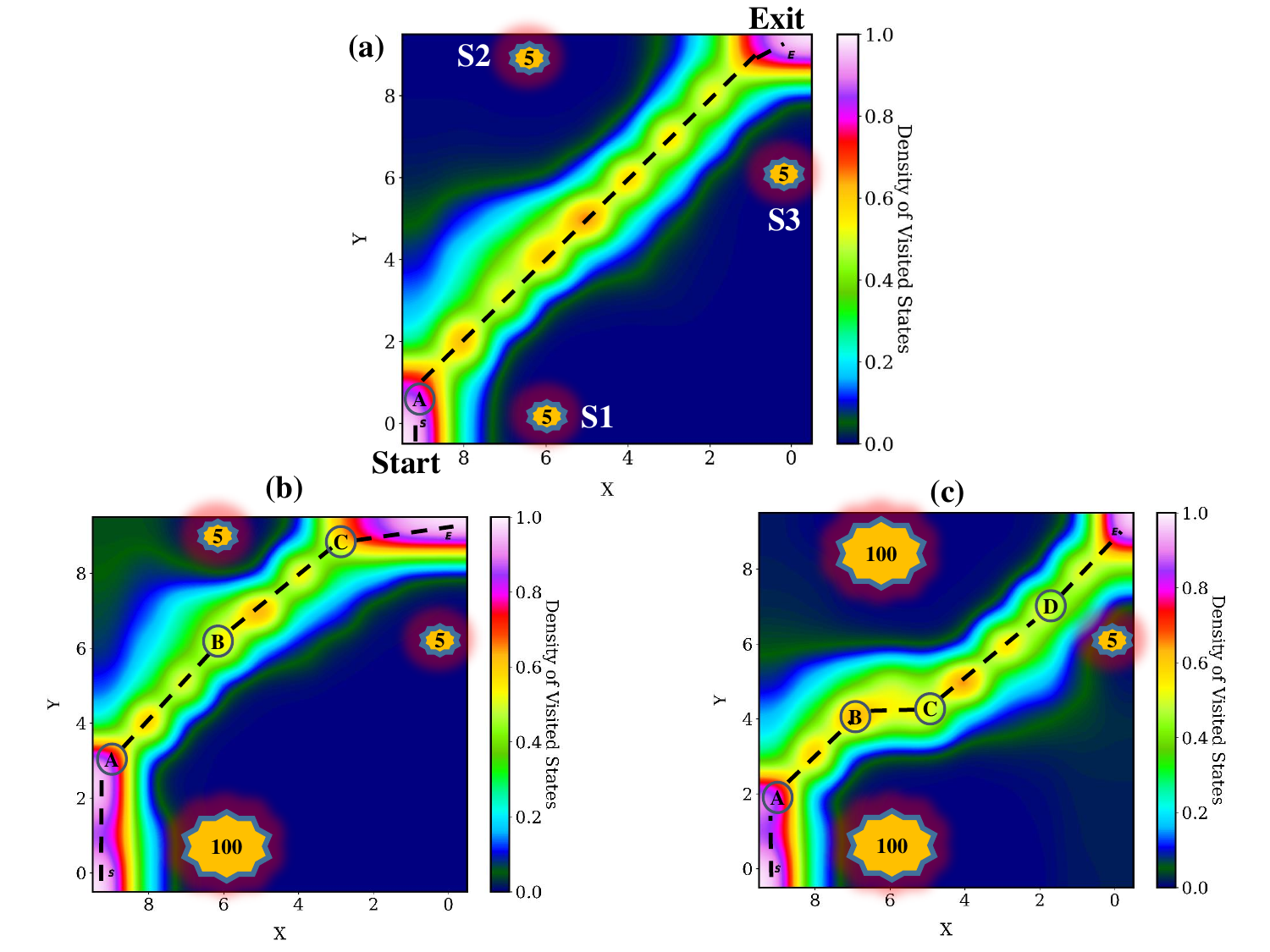}
    \caption{Plot shows the optimum path (black dashed line) predicted by RadDQN in case of three sources (S1, S2 and S3) in simulated floor. The important diversion points within the predicted path in response to the change in radiation intensity of sources are shown as A/B/C/D. (a) Top panel (\textit{Case V1}): S1, S2 and S3 has equal radiation strength (5 unit) (b) bottom left panel (\textit{Case V2}): radiation strength of S1 source is increased by 20-fold. The trajectory has three major diversion points to minimize the radiation exposure. (c) bottom right panel (\textit{Case V3}): radiation strength of S1 and S2 sources are increased by 20-fold. The trajectory has four major diversion points to minimize the radiation exposure.}
    \label{fig:three_sources_strength_compare}
\end{figure*}

So far, we have discussed the scenarios where the variation of number and location of radiation sources leading to change in the optimum path in each cases. However, the radiation strength of sources are considered the same. In realistic environment, it is anticipated to have distribution of radiation hot-spots of unequal strength. Here, we test the sensitivity of RadDQN in such scenarios involving multiple radiation sources with variable radiation strength. Figure \ref{fig:three_sources_strength_compare} depicts such a scenario where three sources S1, S2 and S3 are placed strategically so that the impact of variable  source strength on optimum path becomes evident. In \textbf{Case V1}, all three sources have equal strength. In \textbf{Case V2}, the source strength of S1 is increased by 20-fold as compared to the rest of the radiation sources. Whereas, in \textbf{Case V3}, the strength of S1 and S2 are increased simultaneously by 20-fold as compared to S3. Upon training, RadDQN successfully provided optimum paths for each of these cases. Analysis of the predicted trajectory indicates couple of important insights. In \textbf{Case V2}, to avoid the radiation exposure from S1, agent delayed the diversion toward right (point A) as compared to \textbf{Case V1}. Whereas diversion at point B and C was necessary not only to balance radiation exposure from S2 and S3 but also for reaching exit early. While, for \textbf{Case V3}, higher but equal radiation intensity of S1 and S2 results in two important diversion point (B and C) that allowed agent to equally share the exposure from S1 and S2. At point D, agent avoid exposure from S2 by keeping close to the low intensity radiation field before reaching the exit. Importantly, these changes of trajectories in response to the variation in radiation strength points to the robustness of radiation-aware reward function.

\subsubsection{Performance of RadDQN against vanilla DQN}

Now we provide the detailed comparative analysis on the performance of RadDQN against vanilla DQN approach. This can be considered an ablation study to underpin the effect of each modification that led to the RadDQN architecture. Note that the hyperparameters (Table \ref{tab:t_param}) are kept constant for all the studied cases, expect for case V where the value of parameter \textit{m} that delays the update frequency is taken as 1 instead of 2 (cf. algorithm \ref{alg:sync}).

\begin{table*}[]
\resizebox{\linewidth}{!}
{
\begin{tabular}{@{}c|cccccc@{}}
\toprule
         & exp$_{v}$ \& S$_{v}$ & exp$_{v}$ \& S$_{improv}$ & exp$_{pr}$ \& S$_{v}$ & exp$_{pr}$ \& S$_{improv}$ & exp$_{r}$ \& S$_{v}$ & exp$_{r}$ \& S$_{improv}$ \\ \midrule
Case I   & -0.09          & -0.10               & 0.46            & 0.41                 & \textbf{0.49}           & \textbf{0.49}                \\
Case II  & -0.14          & -0.14               & 0.24            & 0.29                 & 0.41           & \textbf{0.42}                \\
Case III & -0.48          & -0.47               & 0.001           & 0.07                 & \textbf{0.13}           & 0.11                \\
Case IV  & -0.54          & -0.54               & \textbf{-0.16}           & -0.20                & DNC          & DNC               \\
Case V   & -0.42          & -0.44               & 0.07            & 0.02                 & \textbf{0.14}           & 0.13                \\ \bottomrule
\end{tabular}
}
\caption{Average cumulative reward per episode (i.e. sum of discounted rewards per episode after training) subject to variable exploration and update strategy. Exp and S correspond to the exploration and update frequency, respectively. The subscript \textit{v}, \textit{pr}, \textit{r} indicate the vanilla, partially-restricted and restricted exploration strategy, respectively. Subscript \textit{v} and \textit{improv} correspond to vanilla-mode and improvised-mode respectively. The best performing strategies are highlighted in bold. DNC correspond to \textit{did not converge}.}
\footnotesize{}
\label{tab: av_reward_ablation}
\end{table*}

\begin{figure}[htbp!]
    \centering
    \includegraphics[width=\linewidth]{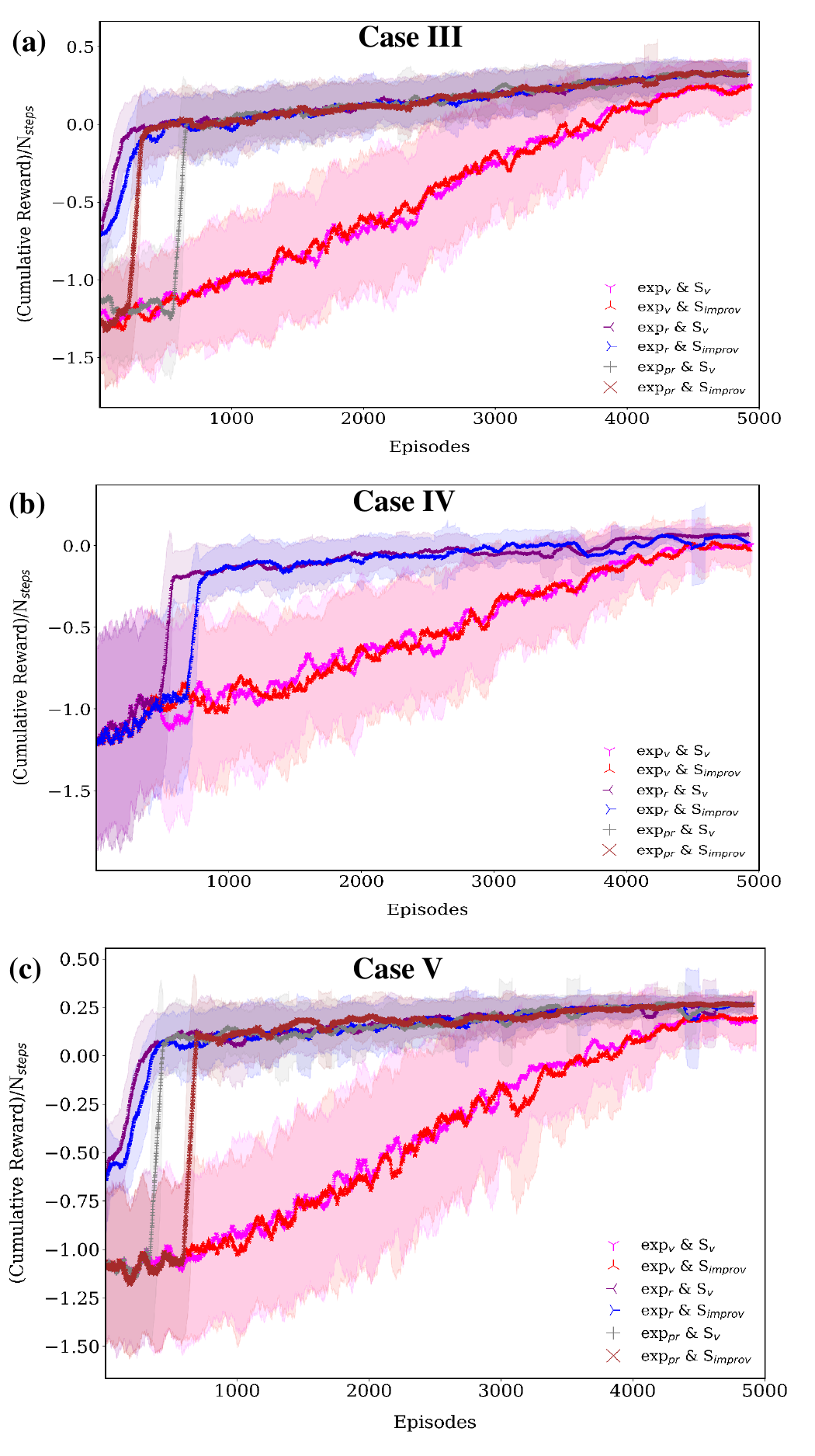}
    \caption{Plot of reward vs. Episode that shows the convergence of training after training subject to the variable exploration and update strategy. (a), (b) and (c) correspond to the studied cases with three sources. Plots for the scenario with two sources are provided in the supporting information.}
    \label{fig:three_sources_training_compare}
\end{figure}

Figure \ref{fig:three_sources_training_compare} depicts the change in average cumulative reward per episode during training. It is desirable that in the course of training the agent would optimize correct set of actions (therefore guide the path) to provide higher cumulative reward with minimum number of steps. vanilla DQN that relies on exploration with $\epsilon$-greedy algorithm evidently suffers from two major issues on achieving the desired objective. Firstly, training with vanilla DQN showed a high moving variance of cumulative reward. Secondly, slow convergence with higher incidence of failure is evident in all the studied cases (Figure \ref{fig:three_sources_reward_compare} and Supporting information, Figure \ref{fig:two_sources_training_compare}, \ref{fig:two_sources_training_compare_winmoves} and \ref{fig:three_sources_training_compare_winmoves}). On the other hand, exploration strategies (\textit{exp$_{pr}$}) and (\textit{exp$_{r}$})) opted in RadDQN observed to quickly find the optimized set of steps that maximizes the cumulative rewards. Further, a significant reduction in the moving variance during training which decays further as the training progress suggesting stable convergence. This is also reflected in Table \ref{tab: av_reward_ablation}, which tabulates the average reward with respect to the total number of training episodes. Furthermore, compared to RadDQN, vanilla DQN failed to converge to the best possible path at the end of the training. The analysis on similarities of the visited trajectories with respect to the ground truth supports the better performance under exp$_{pr}$ with exp$_{r}$ strategy (Supporting information, Figure \ref{fig:frechet_CaseI}-\ref{fig:frechet_CaseV}).

The relative performance of \textit{exp$_{pr}$} against \textit{exp$_{r}$} strategy is noted to be somewhat dependent on the simulation scenario. Compare to \textit{exp$_{pr}$}, a sharp increase in average cumulative reward indicates that the agent is able to learn more quickly under the \textit{exp$_{r}$} strategy (Figure \ref{fig:three_sources_training_compare}). In order to scrutinize the impact of restricting random exploration, we investigate the nature, frequency and distribution of taken actions during training. These actions can be of two types either random (type \textit{r}) or model-directed (i.e. not-random). Again, model-directed actions could originate due to two different conditions as described in algorithm \ref{alg:exploit}, namely (a) actions due to the value of generated random number higher than $\epsilon$ (type \textit{nr}) and (b) actions that are supposed to be random based on $\epsilon$-greedy algorithm but changed to model-directed considering the aspects of future reward and/or winning ratio (type \textit{f/p}). One immediate change that one can anticipate upon activation of exp$_{pr}$ or exp$_{r}$ strategy over the $\epsilon$-greedy exploration is the decrease in the number of random actions. Generated 2D-histograms of executed actions during training confirm this assumption (Supporting information, Figure \ref{fig:random_CaseI}-\ref{fig:random_CaseV}). Importantly, the extent of such changes are predominantly more on those states where radiation exposure is relatively lower as compared to the surrounding states. Upon comparison with results of \textit{exp$_{v}$}, we found that the majority of random actions that converted into model-directed actions lies away from the high radiation field. This permits the model to concentrate more on the states of relatively lower radiation fields, resulting in correction in the weights of NN for the appropriate action (Supporting information, Figure \ref{fig:random_CaseI}-\ref{fig:random_CaseV}). This is observed more acutely under the condition \textit{exp$_{r}$} than \textit{exp$_{pr}$} because the latter imposes stricter conditions based on the future rewards and the winning ratio for the conversion of random action into a model-directed action. However, it is to be noted that under some scenarios the agent may experience short-sightedness as a consequence of the significant reduction in random actions caused by the \textit{exp$_{r}$} strategy. For instance, imagine an environment where the agent occupies a zone where nearby cells have similar radiation intensity (therefore no significant reward in nearby states with respect to the present state). Exploration can be significantly hindered by the \textit{exp$_{r}$} approach in this case, resulting in the agent getting stuck in the so-far explored zone. We observed this behavior in case IV scenario when the agent is trained with \textit{exp$_{r}$}. The agent's short-sightedness during training prevented NN from fine-tuning the weights with respect to the reward table across the entire grid-space. We found that the \textit{exp$_{pr}$} strategy is more effective than \textit{exp$_{r}$} in this type of scenario. The benefit of improvising the update frequency (following algorithm \ref{alg:sync}) depends on the studied environments. Evidently, the improvisation has led to quicker learning in case II and III (Figure \ref{fig:three_sources_training_compare} and \ref{fig:two_sources_training_compare}) as we found steeper rise in cumulative reward as compared to the training with $\epsilon$-greedy exploration. However, opposite behaviour is seen under case V, making the fruit of this implementation environment-sensitive.

Finally, we test the sensitivity of RadDQN on choice of random seed for Case I (i.e. with two sources at (2,2) and (7,7) position). Analysis of convergence showed insignificant changes in training pattern.(Supporting information, Figure \ref{fig:random_seed}) It indicates smooth convergence and training stability irrespective of the choice of seed and therefore, confirms that random seed has minimal dependency on the performance of RadDQN.

\section{Conclusion}

DQN is becoming more widely used to solve complex decision-making problems such as automatic navigation or path optimization in dynamic environments. Radiation protection requires the proper path to be chosen in radiation-contaminated zones to minimize radiation exposure to occupational workers or the public. However, the absence of an efficient reward structure and an effective exploration strategy has prevented it from being applied in radiation protection so far. Here, in this article, we propose a RadDQN architecture that provides a time-efficient minimum radiation exposure based optimal path in radiologically contaminated zone. Within RadDQN, we employ a radiation-aware reward function that's effectively takes into account the important factors like the location and strength of radioactive sources and their proximity to the agent and the destination. Further, we propose a set of unique exploration strategies that transforms a random action into a model-directed one considering the radiation in the future state and on the outcome of the training. We show the effectiveness of these implementations by testing them in multiple scenarios with varying number of source and its strength. Furthermore, we evaluate the predicted optimal path by using Dijkstra, a grid-based deterministic method. Our model is found to achieve superior convergence rate and high training stability as compared to vanilla DQN.

One limitation of our developed model is that the agent is trained in static environment i.e. the location of radiation sources is kept fixed during the training phase. This makes the agent's learned policy subjective to the given environment. How the manipulation of the position of radiation sources during individual training modulates the policy is an intriguing issue and we anticipate to bring on light on this aspect in our future investigations. Further, field experiment with the robots using real-time radiation measurements are presently being planned to test the feasibility of implementation.

\section{Code availability}

The source code of RadDQN is accessible at https://github.com/BiswajitSadhu/RadDQN.

\section*{Acknowledgment}

Authors thank Prof. Y. S. Mayya, IIT Bombay for many fruitful scientific discussions on the project. B.S. and S.A thank Shri Kapildeo Singh (Head, SSS), Dr. M. S. Kulkarni (Head, HPD), Dr. D. K. Aswal (Director, HS \& E Group) for continuous support and encouragement.

\ifCLASSOPTIONcaptionsoff
  \newpage
\fi



%

\bibliography{mybibliography.bib}
\bibliographystyle{IEEEtran}

%

\end{document}


\pagebreak

\listoffigures
\pagebreak

\newpage

\pagenumbering{arabic}

\pagebreak
\clearpage
\begin{figure}[htbp]
    \renewcommand\thefigure{S\arabic{figure}}
    \centering
    \includegraphics[scale=0.6]{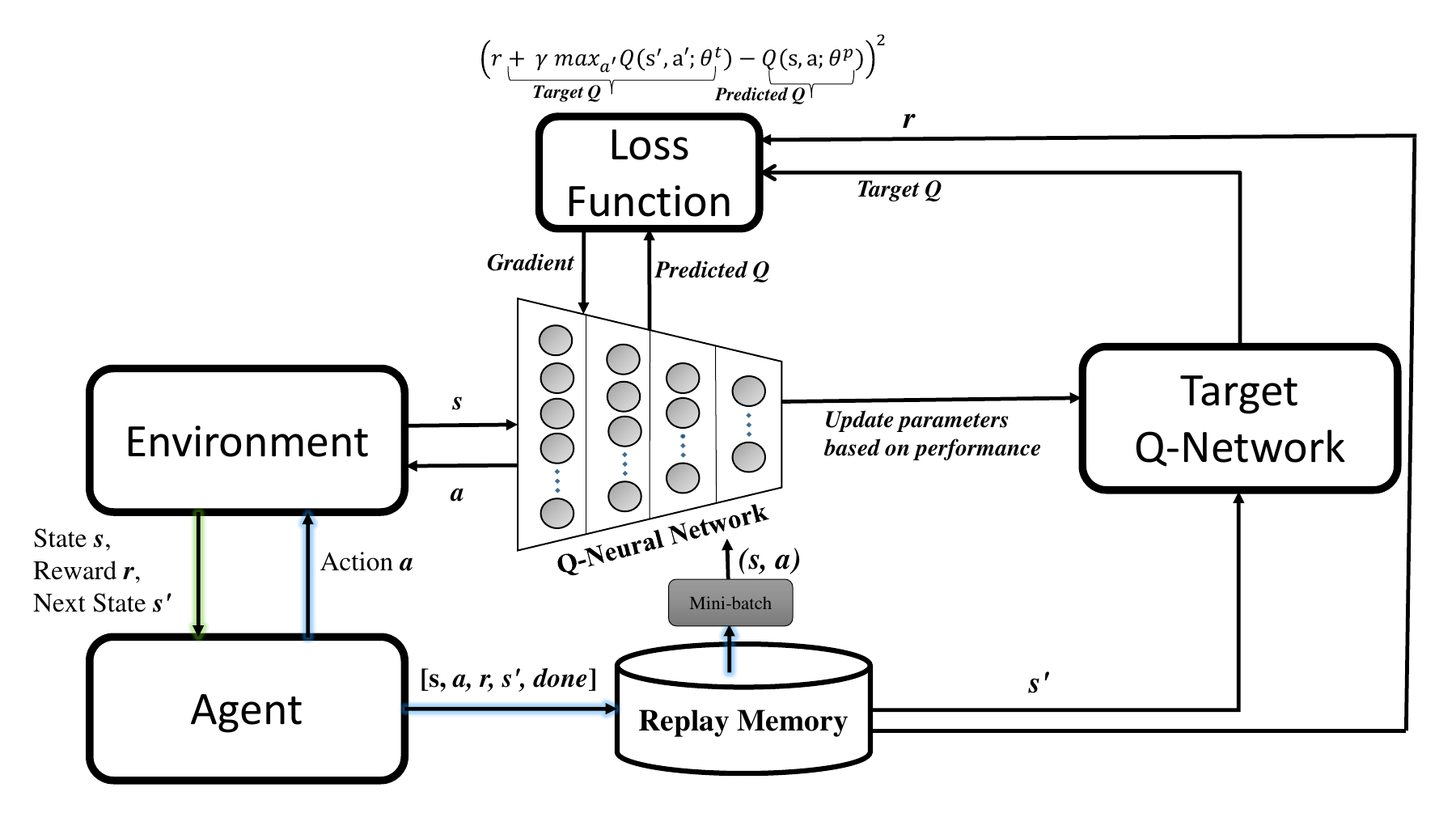}
    \caption{Flowchart of DQN algorithm. $\theta$ in the loss function indicate the trainable parameters of Q-neural network.}
    \label{fig:flowchart_dqn}
\end{figure}

\pagebreak
\clearpage
\begin{figure}[htbp]
    \renewcommand\thefigure{S\arabic{figure}}
    \centering
    \includegraphics[scale=0.50]{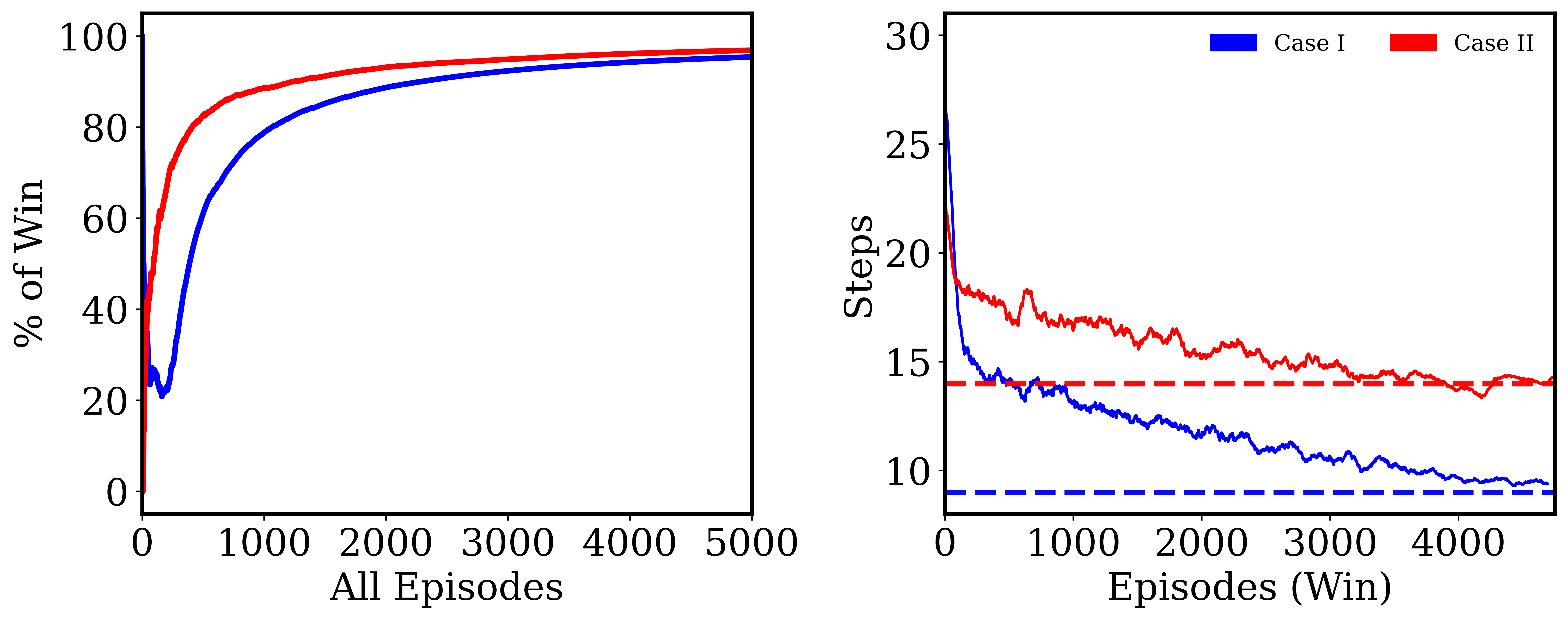} 
    \caption{Left panel: Plot of \% of win vs. number of played episodes. Right panel: Plot of number of steps in the successful episodes vs. number of successful Episodes. The dashed line indicates the ground truth value obtained from grid-based deterministic method.}
    \label{fig:two_sources_winmoves}
\end{figure}

\pagebreak
\clearpage
\begin{figure}[htbp!]
    \renewcommand\thefigure{S\arabic{figure}}
    \centering
    \includegraphics[scale=0.59]{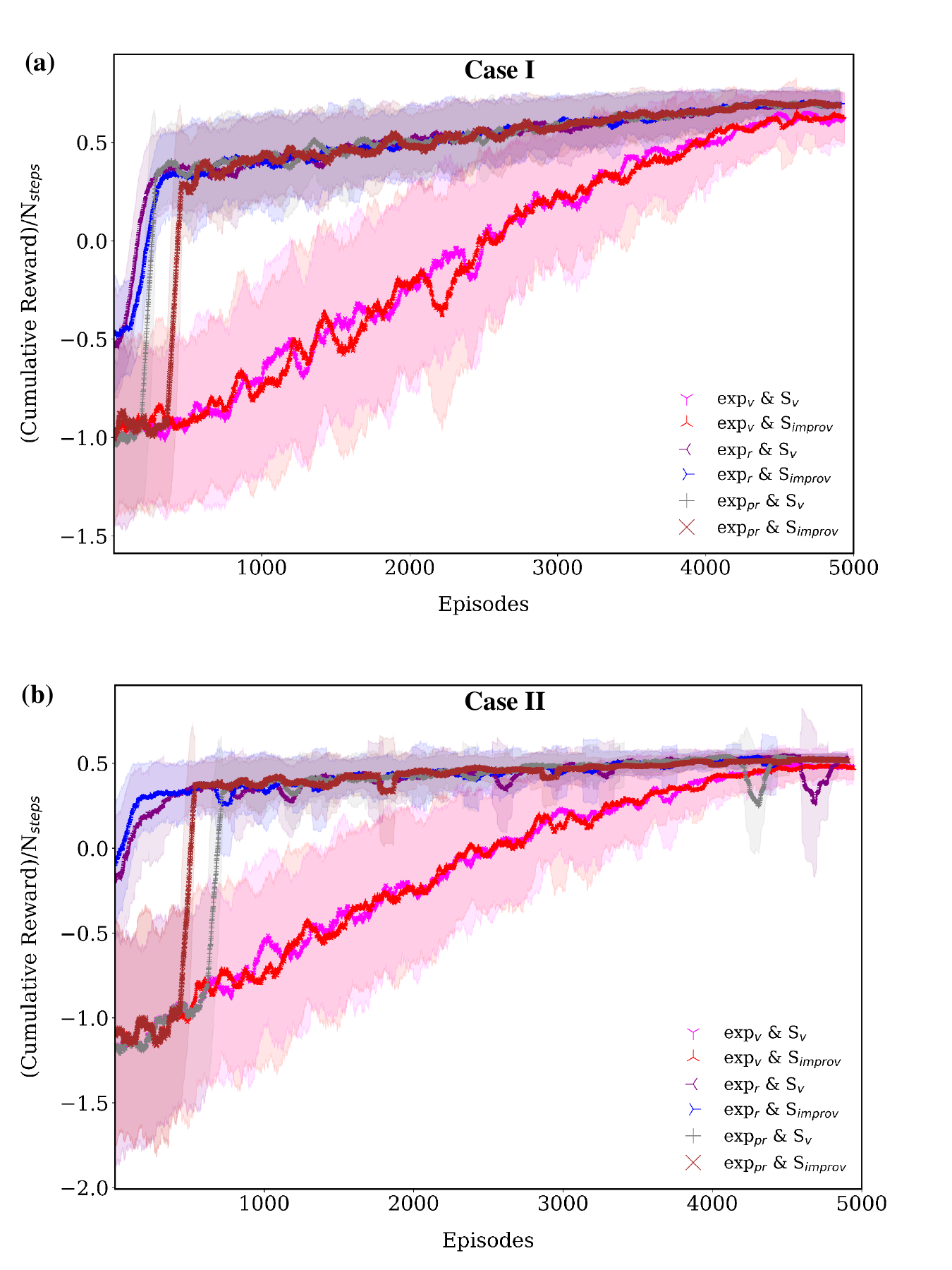}
    \caption{The plots are obtained for scenario with two radioactive sources in floor (Case I and Case II). Plot of average cumulative reward vs. number of played episodes that shows the convergence of training subject to the variable exploration and update strategy.  \textit{exp$_{v}$}, \textit{exp$_{pr}$} and \textit{exp$_{r}$} indicate that the training is performed with vanilla, partially-restricted and restricted exploration strategy, respectively.\textit{s$_{v}$} and \textit{s$_{improv}$} indicates that the update frequency for target Q network during the training is decided via vanilla-mode and improvised-mode, respectively.}
    \label{fig:two_sources_training_compare}
\end{figure}

\pagebreak
\clearpage
\begin{figure}[htbp!]
    \renewcommand\thefigure{S\arabic{figure}}
    \centering
    \includegraphics[scale=0.59]{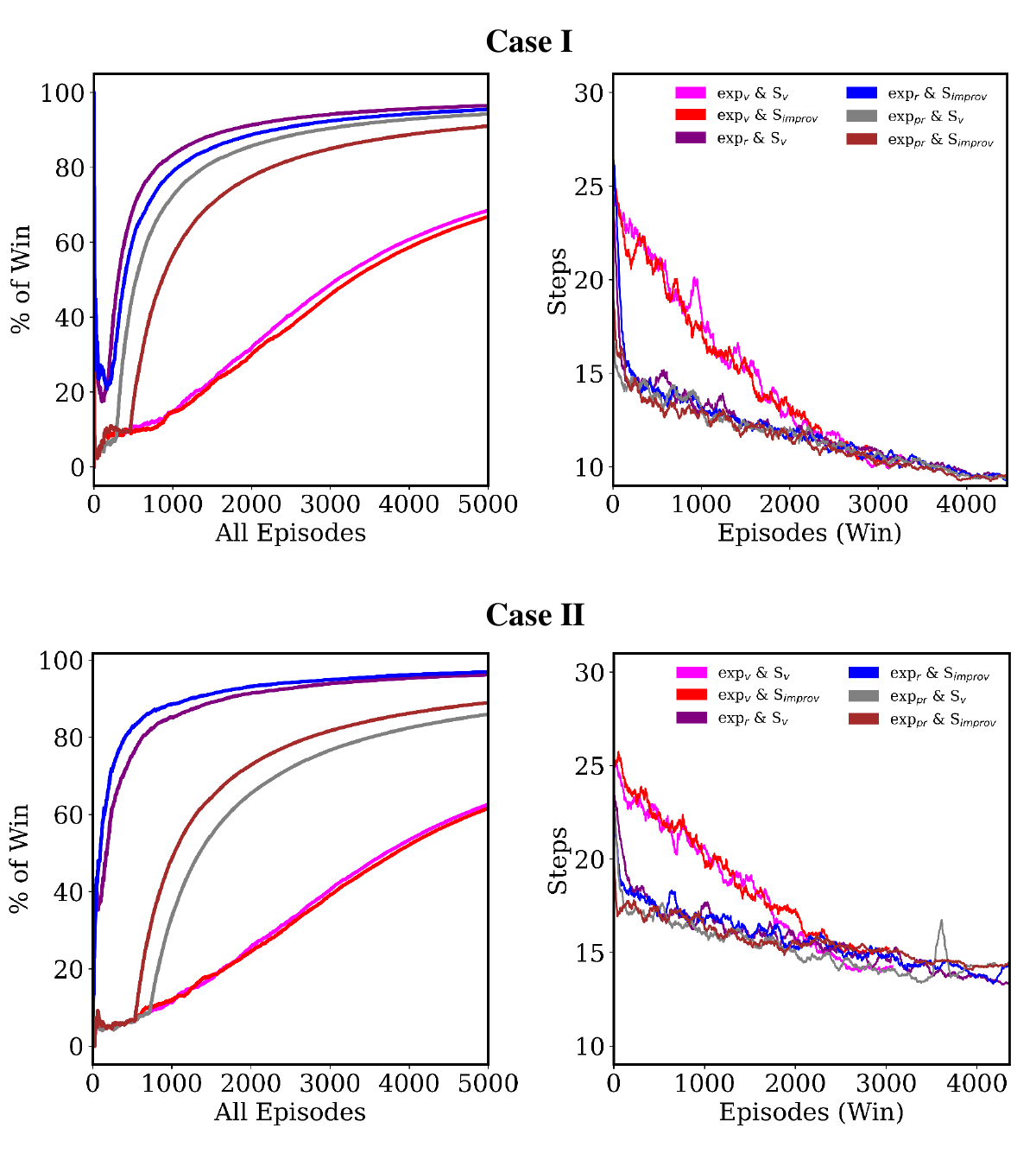}
    \caption{The plots are obtained for scenario with two radioactive sources in floor (Case I and Case II). Left panel: Plot of \% of win vs. number of played episodes. Right panel: Plot of number of steps in the successful episodes vs. number of successful Episodes. \textit{exp$_{v}$}, \textit{exp$_{pr}$} and \textit{exp$_{r}$} indicate that the training is performed with vanilla, partially-restricted and restricted exploration strategy, respectively.\textit{s$_{v}$} and \textit{s$_{improv}$} indicates that the update frequency for target Q network during the training is decided via vanilla-mode and improvised-mode, respectively}
    \label{fig:two_sources_training_compare_winmoves}
\end{figure}

\pagebreak
\clearpage
\begin{figure}[htbp!]
    \renewcommand\thefigure{S\arabic{figure}}
    \centering
    \includegraphics[scale=0.59]{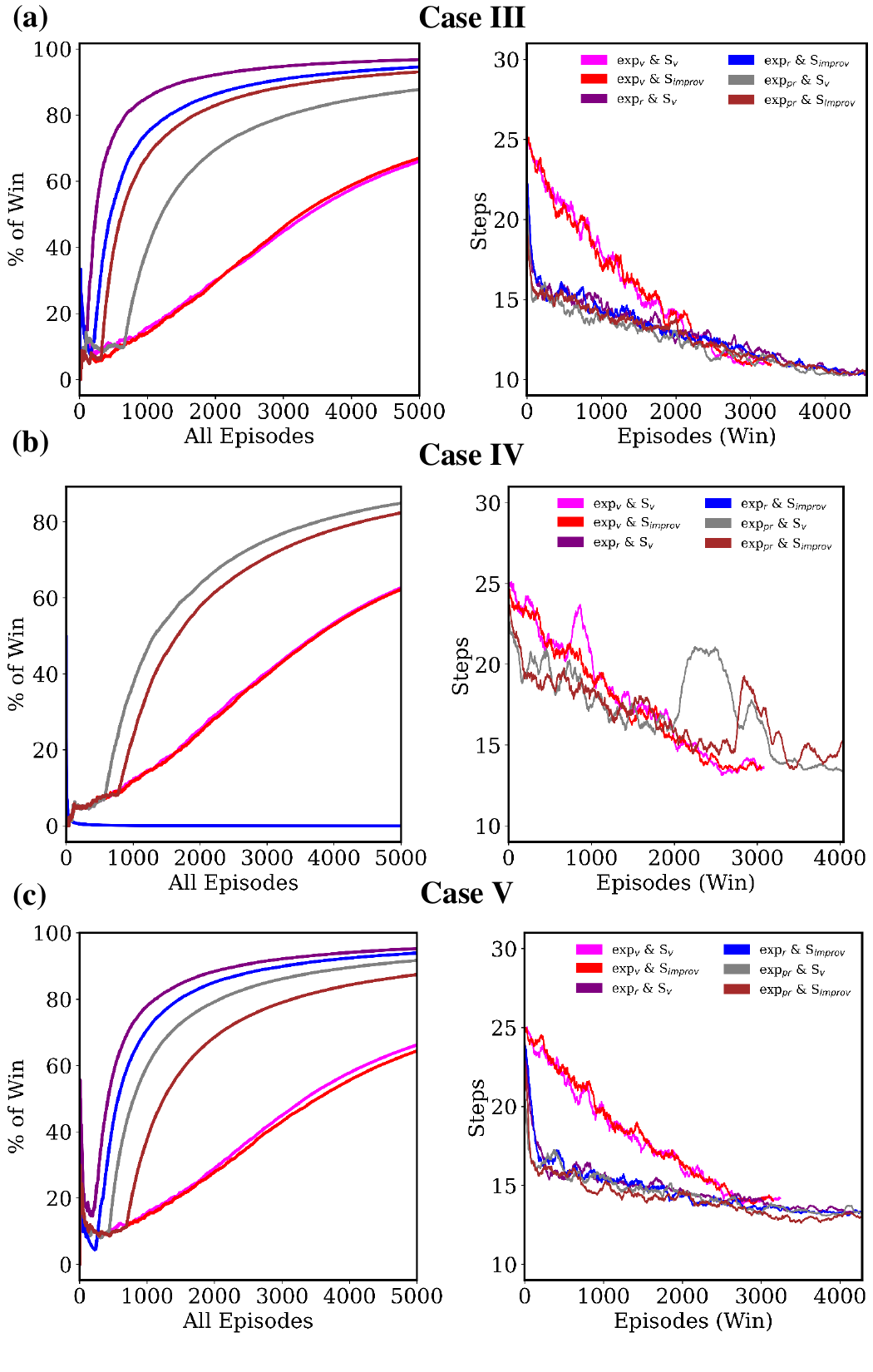}
    \caption{The plots are obtained for scenario with three radioactive sources in floor (Case III, Case IV and Case V)Left panel: Plot of \% of win vs. number of played episodes. Right panel: Plot of number of steps in the successful episodes vs. number of successful Episodes. \textit{exp$_{v}$}, \textit{exp$_{pr}$} and \textit{exp$_{r}$} indicate that the training is performed with vanilla, partially-restricted and restricted exploration strategy, respectively.\textit{s$_{v}$} and \textit{s$_{improv}$} indicates that the update frequency for target Q network during the training is decided via vanilla-mode and improvised-mode, respectively.}
    \label{fig:three_sources_training_compare_winmoves}
\end{figure}

\pagebreak
\clearpage
\begin{figure}[htbp!]
    \renewcommand\thefigure{S\arabic{figure}}
    \centering
    \includegraphics[scale=0.45]{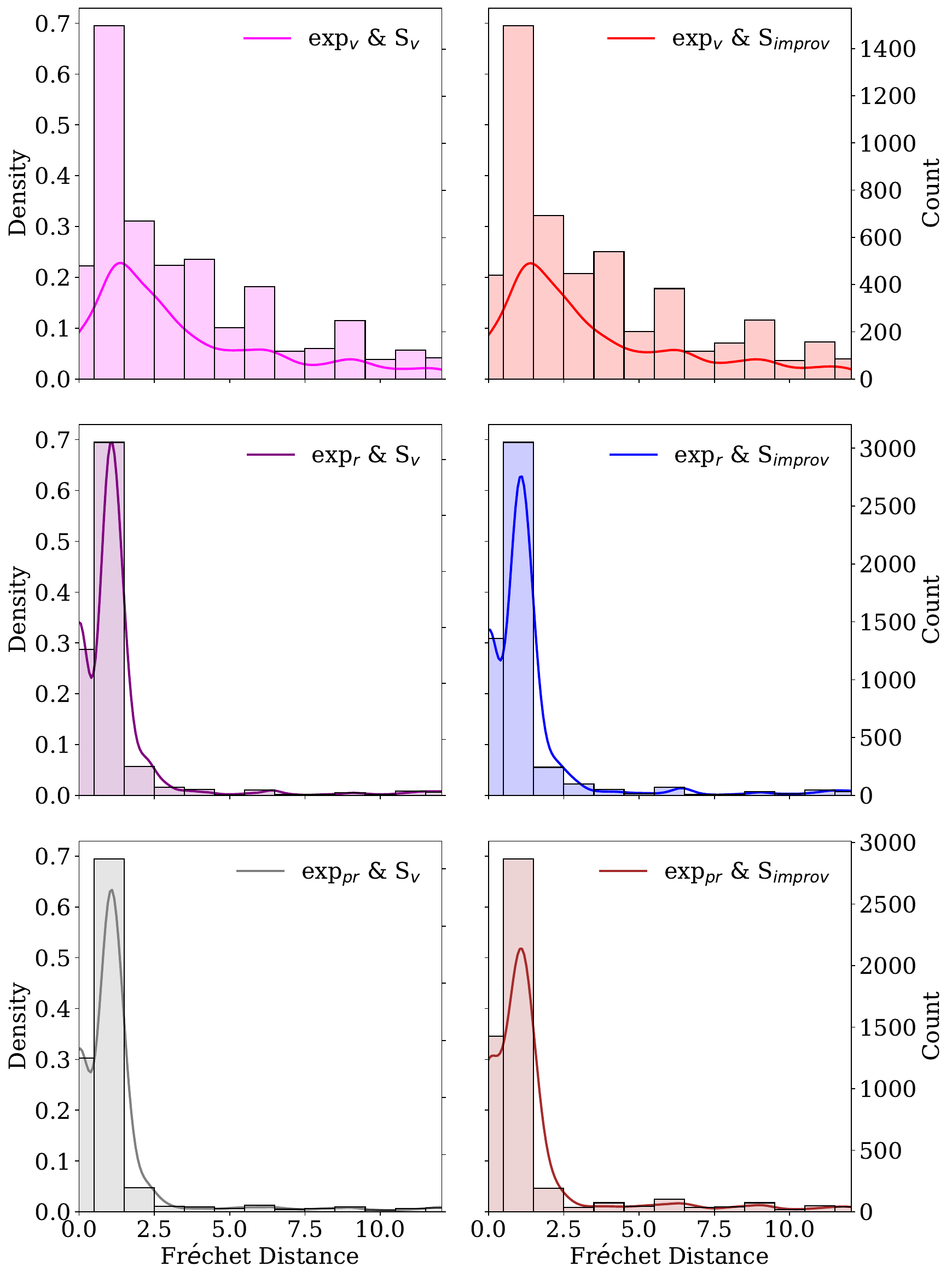}
    \caption{Case-I (two radioactive sources in simulated floor): Comparison of RadDQN predicted path-ensemble with Ground truth. The x axis of the histogram is the Frechet distance between the RadDQN predicted individual trajectories with Dijkstra's algorithm derived path. Whereas, y (twin-y) indicates frequency of trajectory(count) in the path-ensemble. The top, middle and bottom rows shows the plot under \textit{exp$_{v}$}, \textit{exp$_{r}$} and \textit{exp$_{pr}$} strategy, respectively. The figures on the right-side of each row correspond to the cases where algorithm 2 is applied for the computation of the update frequency during the training. The histogram is further fitted to obtain kernel-density estimated plot.}
    \label{fig:frechet_CaseI}
\end{figure}

\pagebreak
\clearpage
\begin{figure}[htbp!]
    \renewcommand\thefigure{S\arabic{figure}}
    \centering
    \includegraphics[scale=0.45]{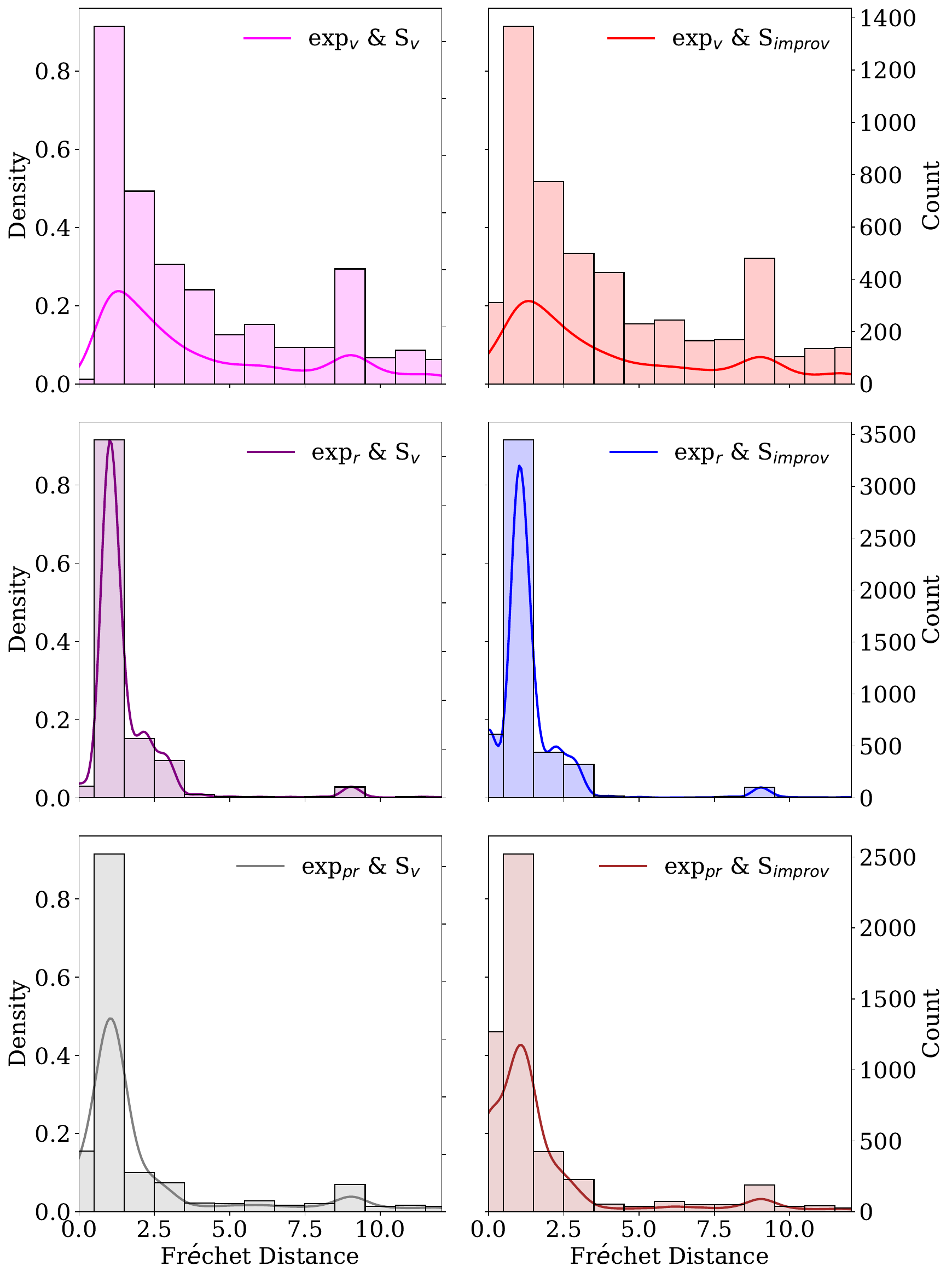}
    \caption{Case-II (two radioactive sources in simulated floor): Comparison of RadDQN predicted path-ensemble with Ground truth. The x axis of the histogram is the Frechet distance between the RadDQN predicted individual trajectories with Dijkstra's algorithm derived path. Whereas, y (twin-y) indicates frequency of trajectory(count) in the path-ensemble. The top, middle and bottom rows shows the plot under \textit{exp$_{v}$}, \textit{exp$_{r}$} and \textit{exp$_{pr}$} strategy, respectively. The figures on the right-side of each row correspond to the cases where algorithm 2 is applied for the computation of the update frequency during the training. The histogram is further fitted to obtain kernel-density estimated plot.}
    \label{fig:frechet_CaseII}
\end{figure}

\pagebreak
\clearpage
\begin{figure}[htbp!]
    \renewcommand\thefigure{S\arabic{figure}}
    \centering
    \includegraphics[scale=0.45]{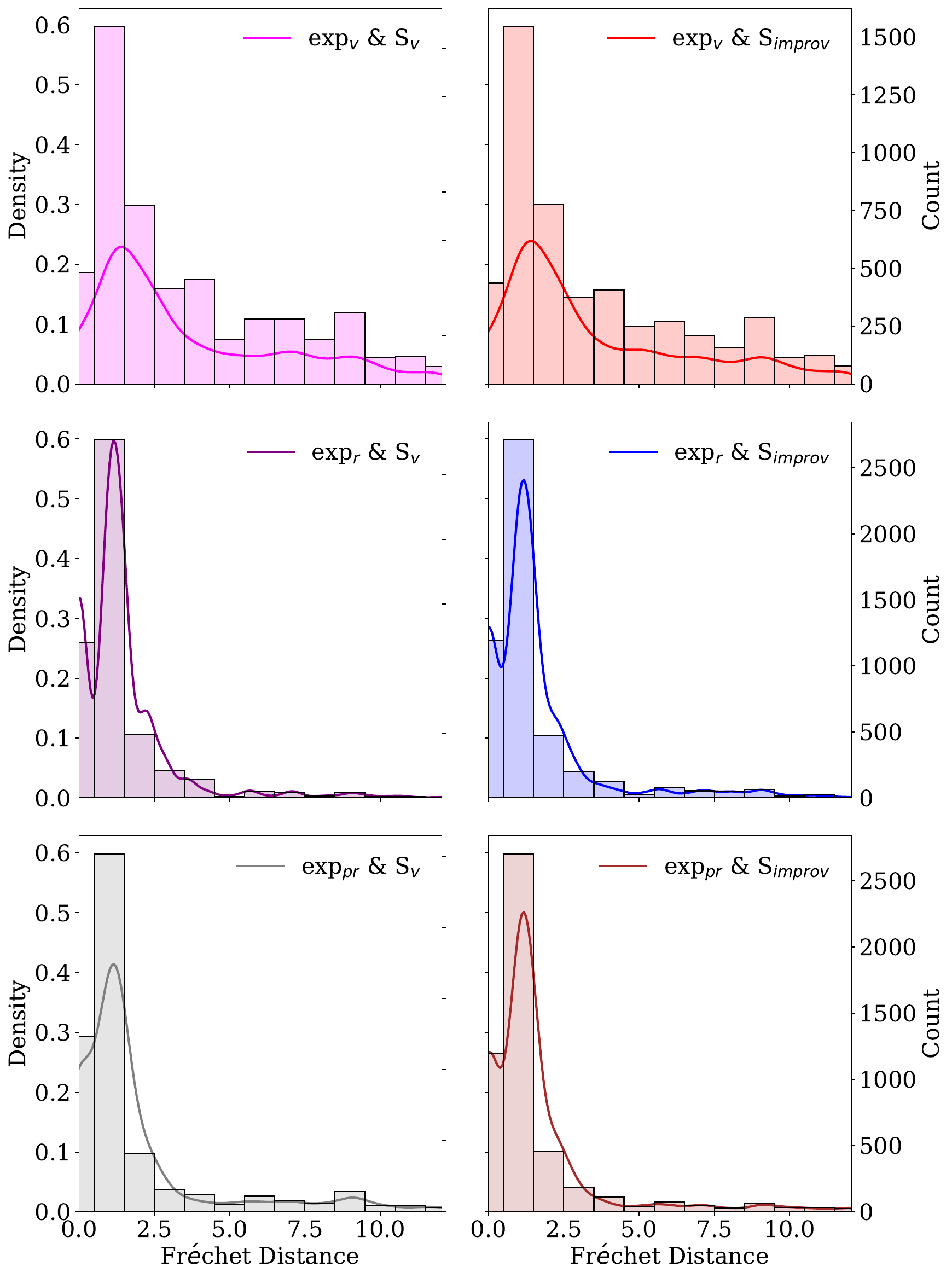}
    \caption{Case-III (three radioactive sources in simulated floor): Comparison of RadDQN predicted path-ensemble with Ground truth. The x axis of the histogram is the Fr\'echet distance between the RadDQN predicted individual trajectories with Dijkstra's algorithm derived path. Whereas, y (twin-y) indicates frequency of trajectory(count) in the path-ensemble. The top, middle and bottom rows shows the plot under \textit{exp$_{v}$}, \textit{exp$_{r}$} and \textit{exp$_{pr}$} strategy, respectively. The figures on the right-side of each row correspond to the cases where algorithm 2 is applied for the computation of the update frequency during the training. The histogram is further fitted to obtain kernel-density estimated plot.}
    \label{fig:frechet_CaseIII}

\end{figure}

\pagebreak
\clearpage
\begin{figure}[htbp!]
    \renewcommand\thefigure{S\arabic{figure}}
    \centering
    \includegraphics[scale=0.45]{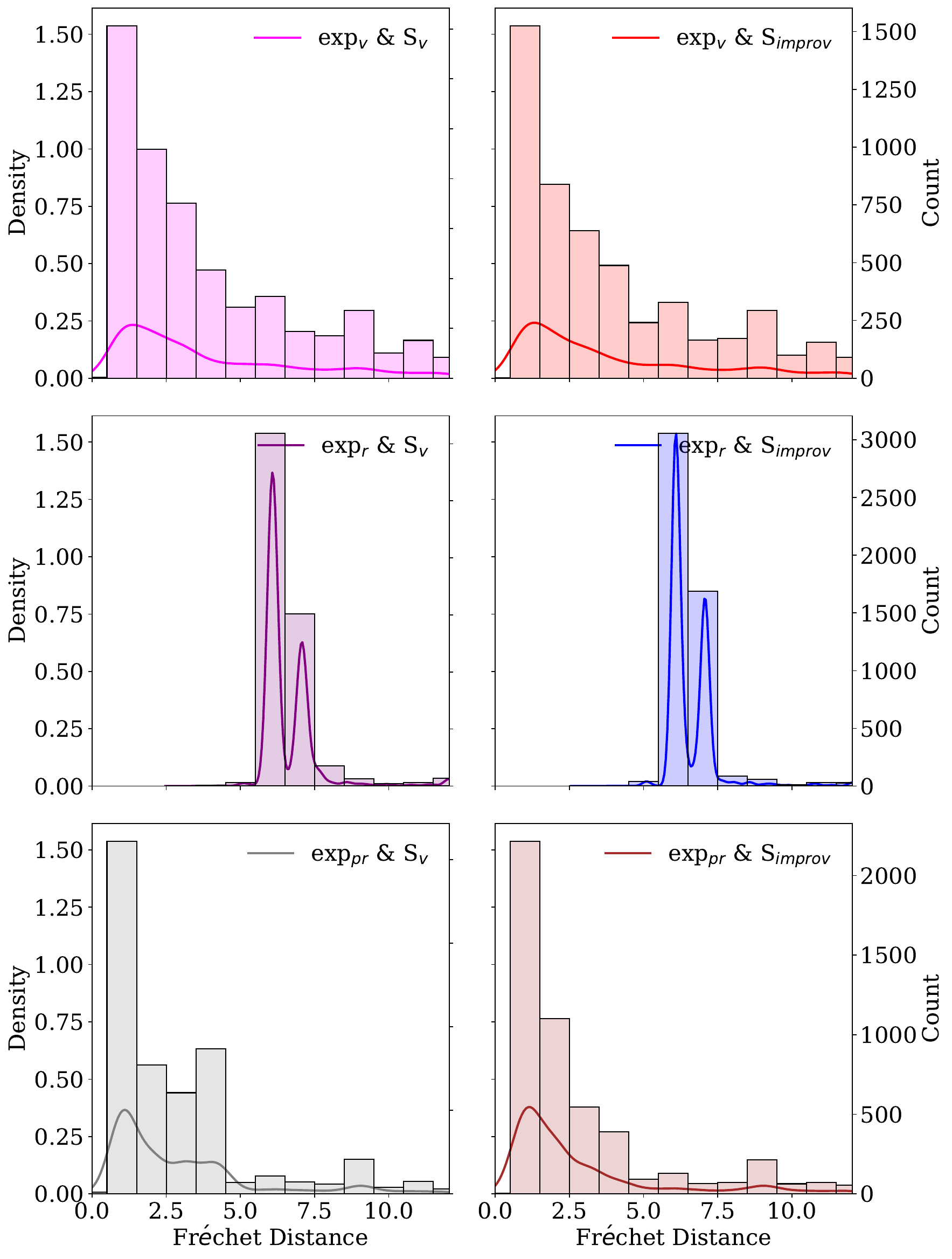}
    \caption{Case-IV (three radioactive sources in simulated floor): Comparison of RadDQN predicted path-ensemble with Ground truth. The x axis of the histogram is the Frechet distance between the RadDQN predicted individual trajectories with Dijkstra's algorithm derived path. Whereas, y (twin-y) indicates frequency of trajectory(count) in the path-ensemble. The top, middle and bottom rows shows the plot under \textit{exp$_{v}$}, \textit{exp$_{r}$} and \textit{exp$_{pr}$} strategy, respectively. The figures on the right-side of each row correspond to the cases where algorithm 2 is applied for the computation of the update frequency during the training. The histogram is further fitted to obtain kernel-density estimated plot. The plots in middle row correspond to unconverged training under \textit{exp$_{r}$} strategy.}
    \label{fig:frechet_CaseIV}
\end{figure}

\pagebreak
\clearpage
\begin{figure}[htbp!]
    \renewcommand\thefigure{S\arabic{figure}}
    \centering
    \includegraphics[scale=0.45]{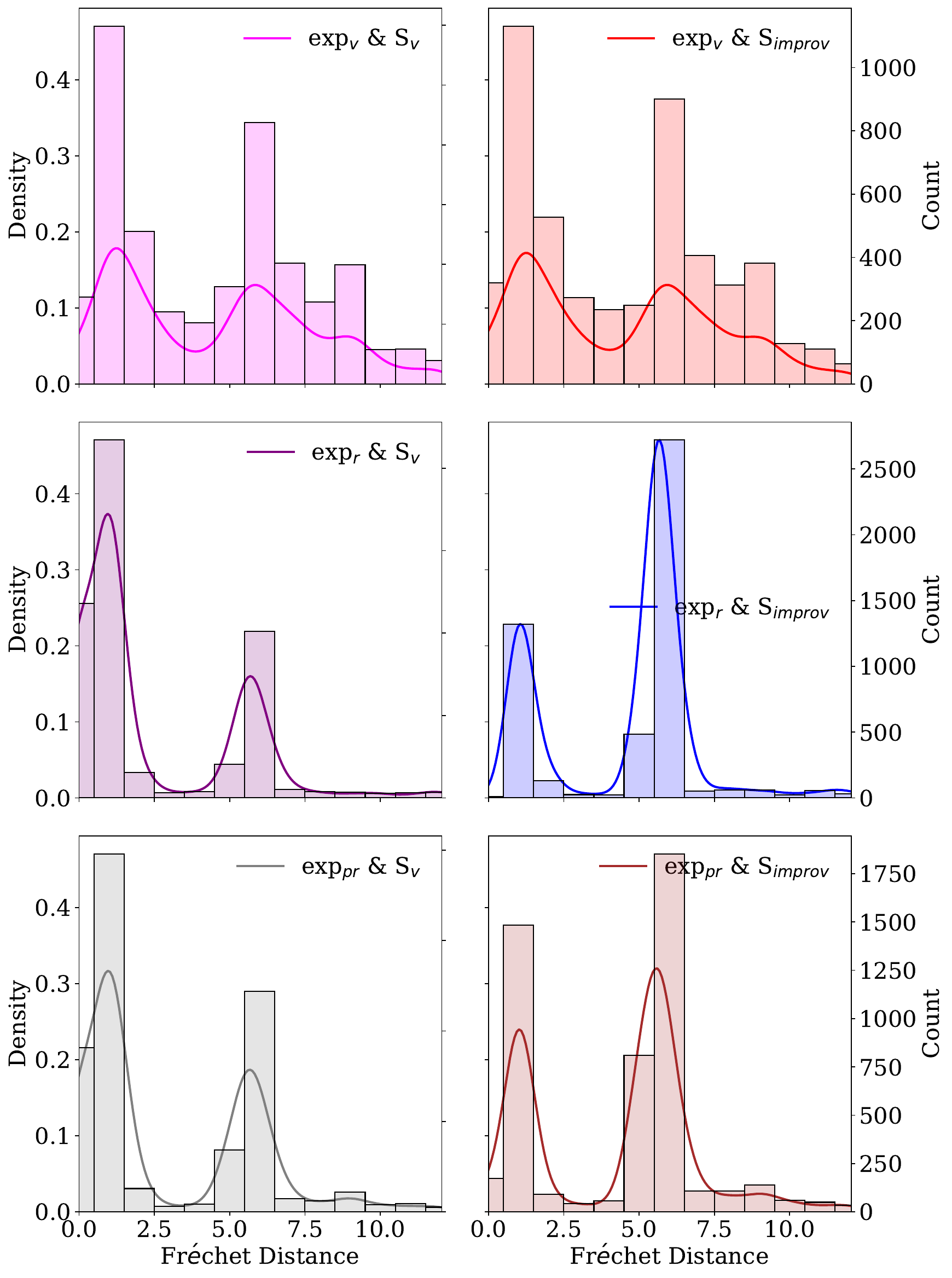}
    \caption{Case-V (three radioactive sources in simulated floor): Comparison of RadDQN predicted path-ensemble with Ground truth. The x axis of the histogram is the Frechet distance between the RadDQN predicted individual trajectories with Dijkstra's algorithm derived path. Whereas, y (twin-y) indicates frequency of trajectory(count) in the path-ensemble. The top, middle and bottom rows shows the plot under \textit{exp$_{v}$}, \textit{exp$_{r}$} and \textit{exp$_{pr}$} strategy, respectively. The figures on the right-side of each row correspond to the cases where algorithm 2 is applied for the computation of the update frequency during the training. The histogram is further fitted to obtain kernel-density estimated (KDE) plot. The double hump observed in the histogram (and kde plot) indicated two possible optimal paths. The Frechet distance is large for one of the hump as the Dijkstra's algorithm based ground truth only predict one path (and ignore others) by definition.}
    \label{fig:frechet_CaseV}
\end{figure}

\pagebreak
\clearpage
\begin{figure}[htbp!]
    \renewcommand\thefigure{S\arabic{figure}}
    \centering
    \includegraphics[scale=0.45]{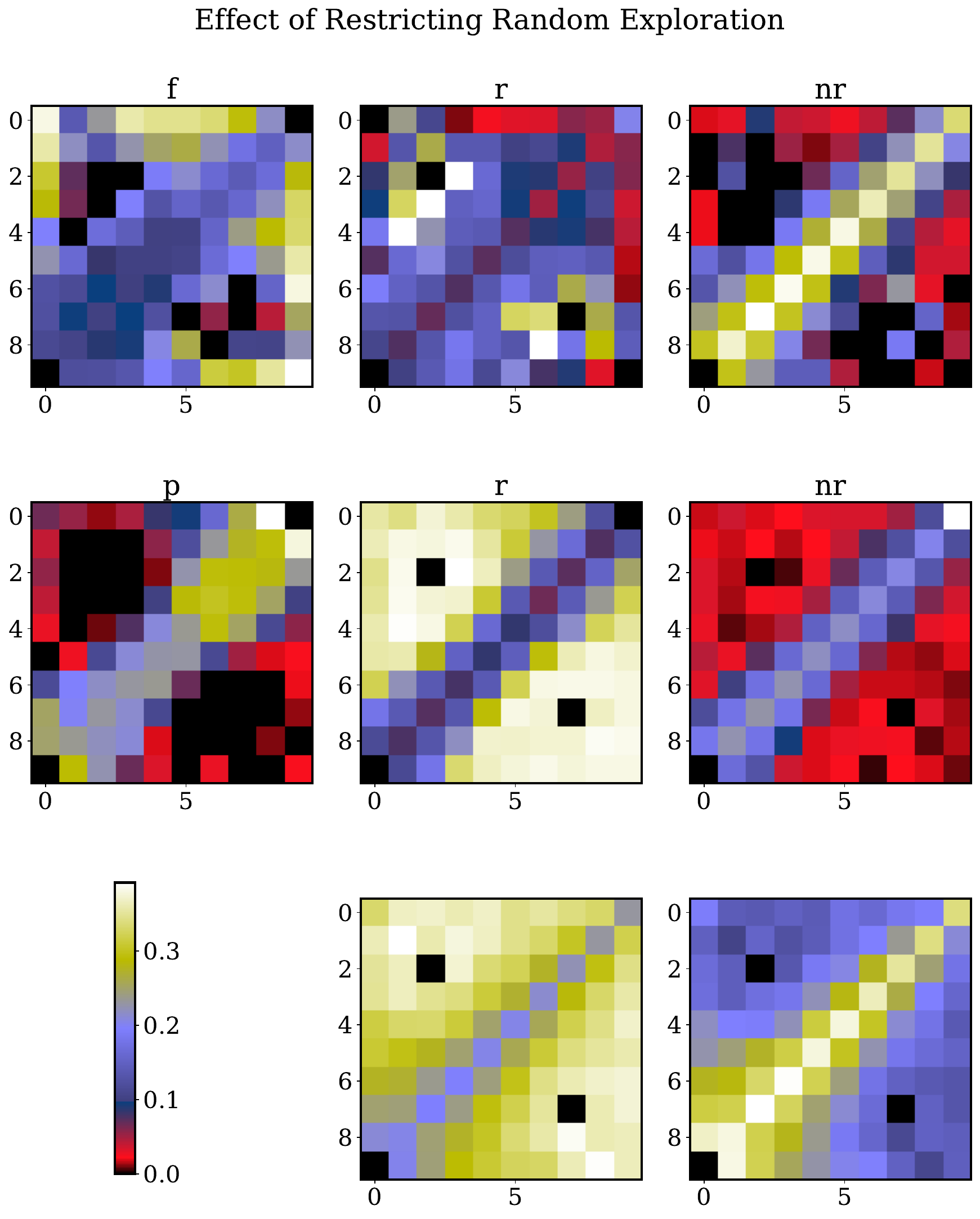}
    \caption{Effect of restricting random exploration in Case I (two radioactive sources in simulated floor):  The nature of action is abbreviated as \textit{r} (i.e. random), \textit{nr} (i.e. not-random or model-directed following $\epsilon$-greedy algorithm) and '\textit{p/f}' (actions that are converted to model-directed following algorithm 1 under \textit{exp$_{pr}$} and \textit{exp$_{r}$} strategy, respectively). The top, middle and bottom rows shows the plot for $exp_{pr}$, $exp_{r}$ and $exp_{v}$ strategy, respectively. The color-bar indicate the state-specific density of action in training.}
    \label{fig:random_CaseI}
\end{figure}

\pagebreak
\clearpage
\begin{figure}[htbp!]
    \renewcommand\thefigure{S\arabic{figure}}
    \centering
    \includegraphics[scale=0.45]{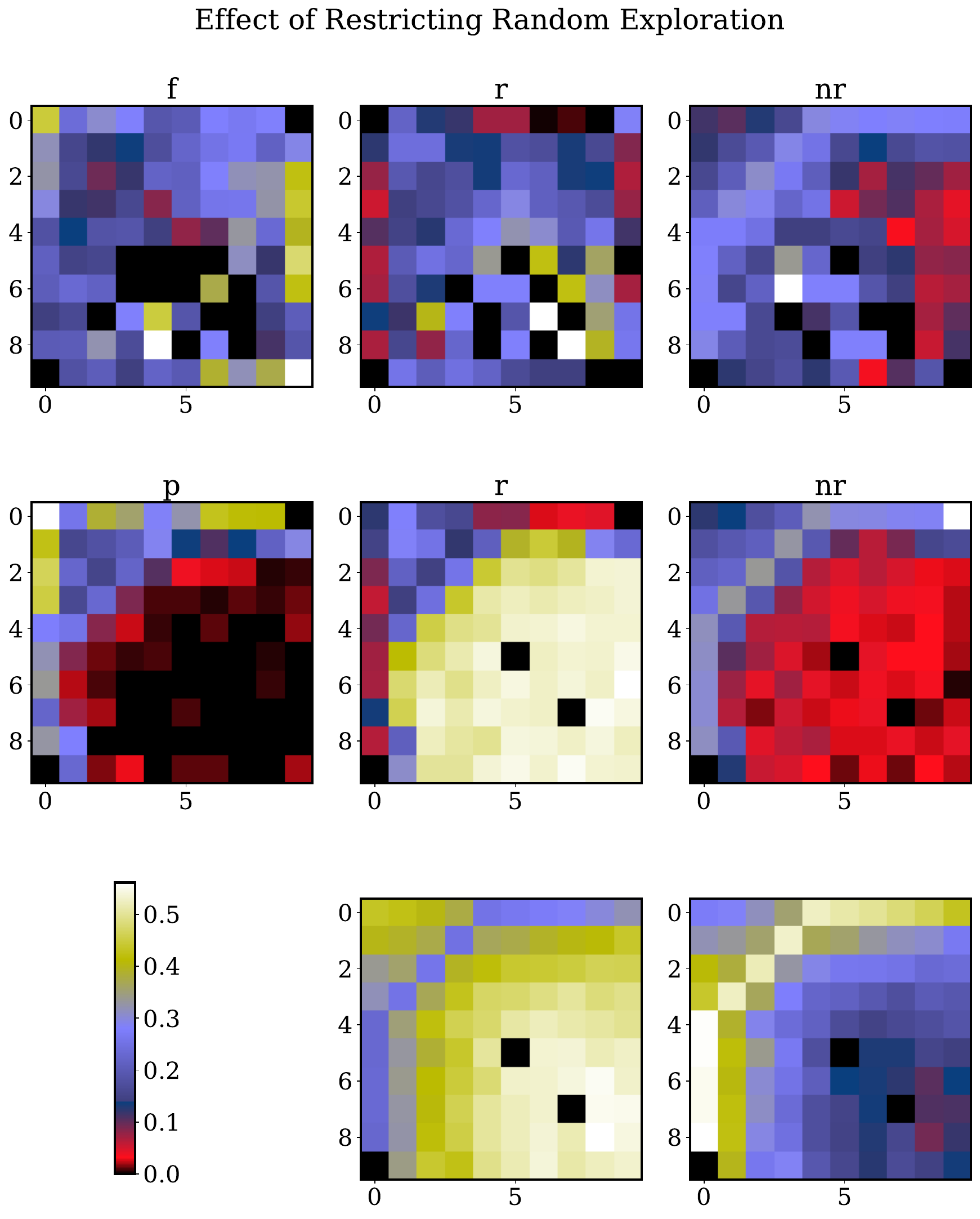}
    \caption{Case-II (two radioactive sources in simulated floor): Effect of restricting random exploration. The nature of action is abbreviated as \textit{r} (i.e. random), \textit{nr} (i.e. not-random or model-directed following $\epsilon$-greedy algorithm) and '\textit{p/f}' (actions that are converted to model-directed following algorithm 1 under \textit{exp$_{pr}$} and \textit{exp$_{r}$} strategy, respectively). The top, middle and bottom rows shows the plot for $exp_{pr}$, $exp_{r}$ and $exp_{v}$ strategy, respectively. The color-bar indicate the state-specific density of action in training.}
    \label{fig:random_CaseII}
\end{figure}

\pagebreak
\clearpage
\begin{figure}[htbp!]
    \renewcommand\thefigure{S\arabic{figure}}
    \centering
    \includegraphics[scale=0.45]{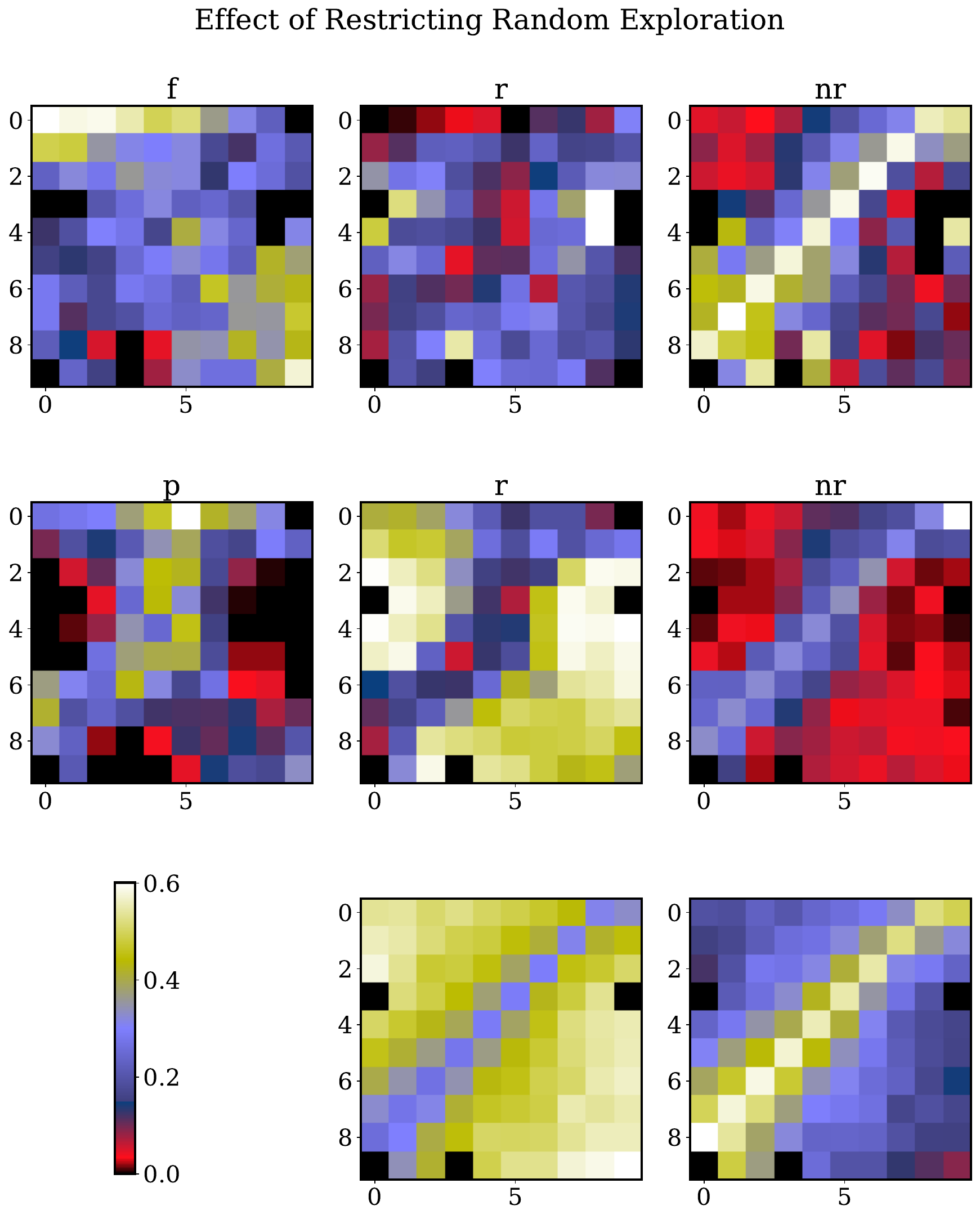}
    \caption{Case-III (three radioactive sources in simulated floor): Effect of restricting random exploration. The nature of action is abbreviated as \textit{r} (i.e. random), \textit{nr} (i.e. not-random or model-directed following $\epsilon$-greedy algorithm) and '\textit{p/f}' (actions that are converted to model-directed following algorithm 1 under \textit{exp$_{pr}$} and \textit{exp$_{r}$} strategy, respectively). The top, middle and bottom rows shows the plot for $exp_{pr}$, $exp_{r}$ and $exp_{v}$ strategy, respectively. The color-bar indicate the state-specific density of action in training.}
    \label{fig:random_CaseIII}
\end{figure}

\pagebreak
\clearpage
\begin{figure}[htbp!]
    \renewcommand\thefigure{S\arabic{figure}}
    \centering
    \includegraphics[scale=0.45]{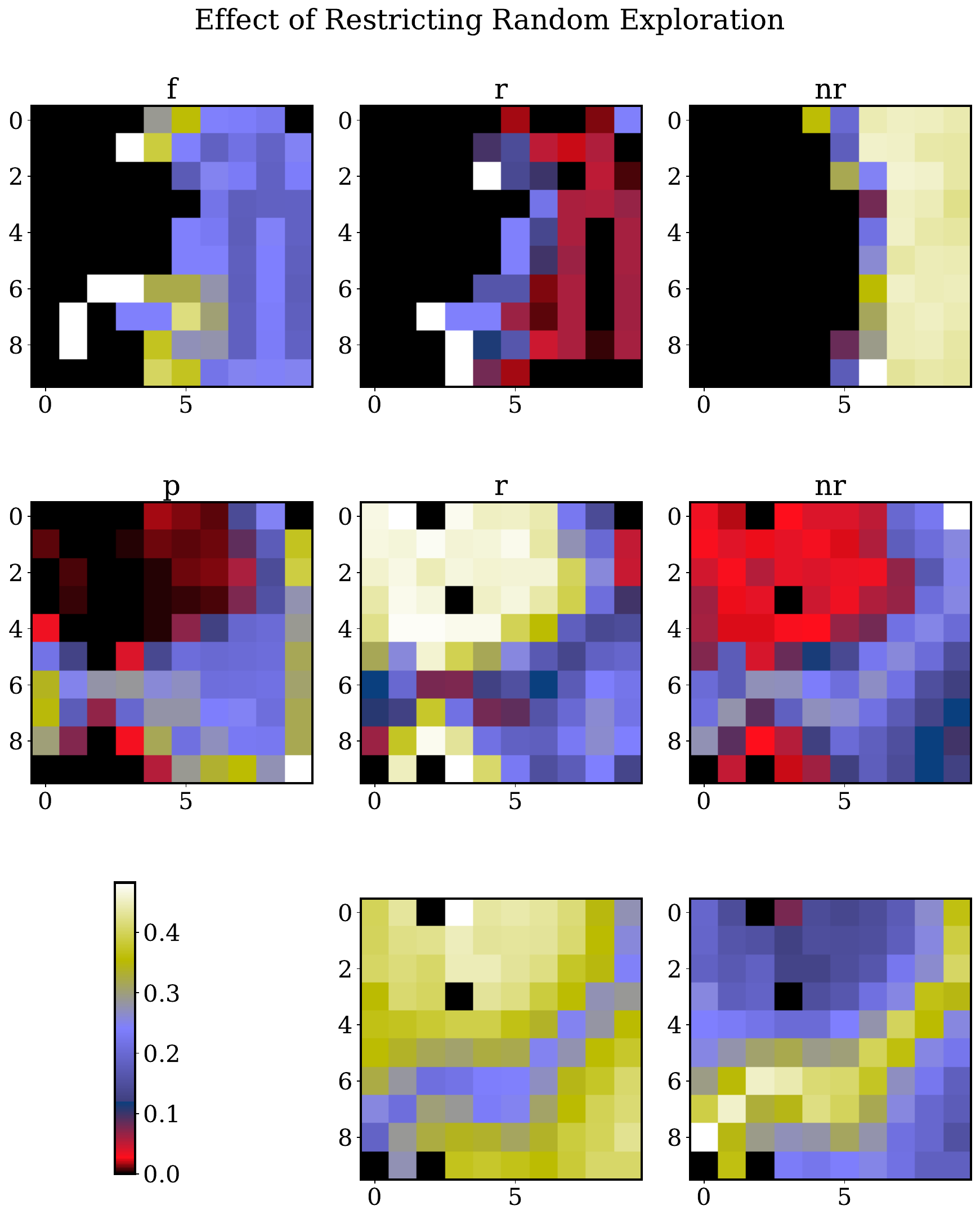}
    \caption{Case-IV (three radioactive sources in simulated floor): Effect of restricting random exploration. The nature of action is abbreviated as \textit{r} (i.e. random), \textit{nr} (i.e. not-random or model-directed following $\epsilon$-greedy algorithm) and '\textit{p/f}' (actions that are converted to model-directed following algorithm 1 under \textit{exp$_{pr}$} and \textit{exp$_{r}$} strategy, respectively). The top, middle and bottom rows shows the plot for $exp_{pr}$, $exp_{r}$ and $exp_{v}$ strategy, respectively. The color-bar indicate the state-specific density of action in training.}
    \label{fig:random_CaseIV}
\end{figure}

\pagebreak
\clearpage
\begin{figure}[htbp!]
    \renewcommand\thefigure{S\arabic{figure}}
    \centering
    \includegraphics[scale=0.45]{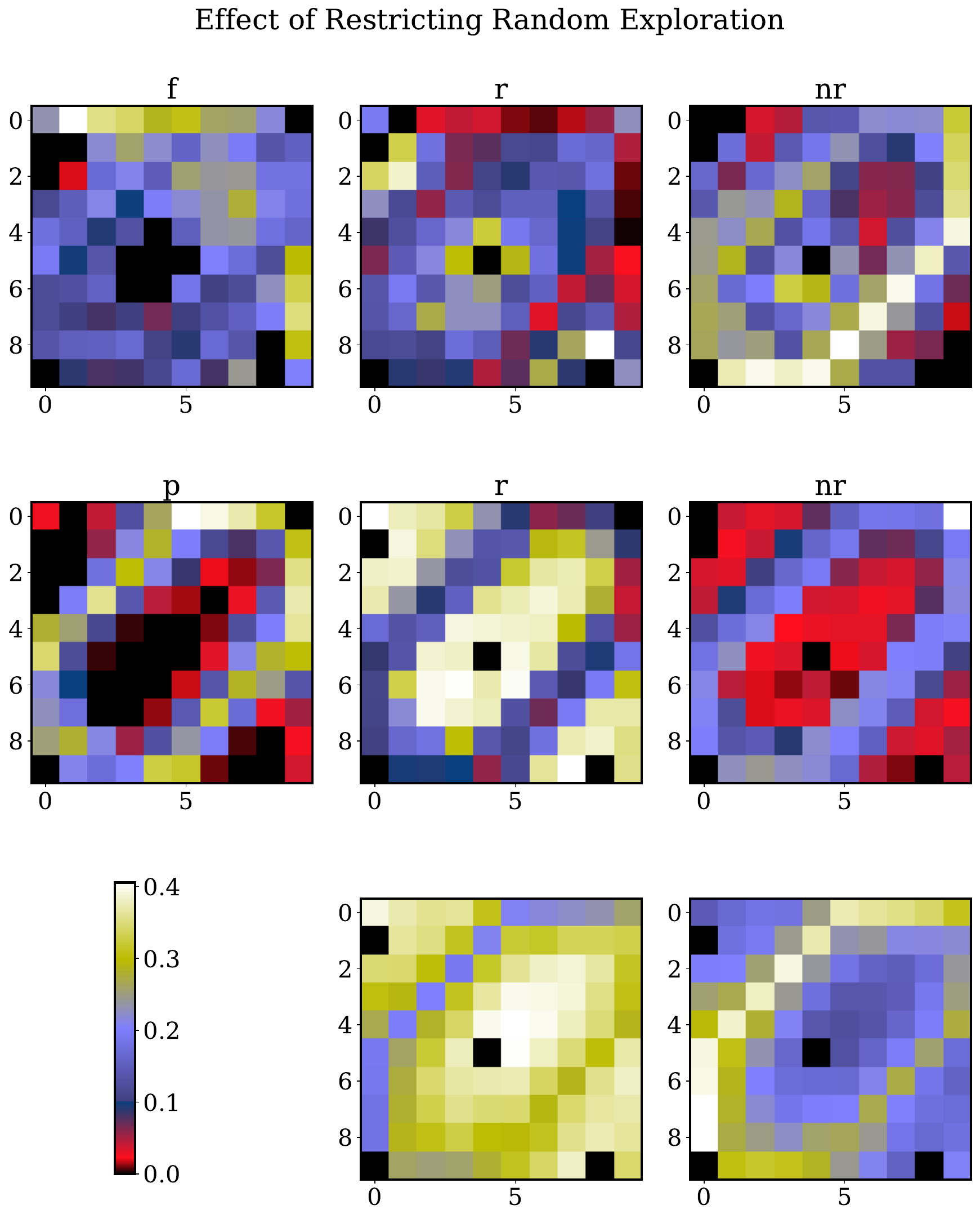}
    \caption{Case-V (three radioactive sources in simulated floor): Effect of restricting random exploration. The nature of action is abbreviated as \textit{r} (i.e. random), \textit{nr} (i.e. not-random or model-directed following $\epsilon$-greedy algorithm) and '\textit{p/f}' (actions that are converted to model-directed following algorithm 1 under \textit{exp$_{pr}$} and \textit{exp$_{r}$} strategy, respectively). The top, middle and bottom rows shows the plot for \textit{exp$_{v}$}, \textit{exp$_{r}$} and \textit{exp$_{pr}$} strategy, respectively. The color-bar indicate the state-specific density of action in training.}
    \label{fig:random_CaseV}
\end{figure}

\pagebreak
\clearpage
\begin{figure}[htbp]
    \renewcommand\thefigure{S\arabic{figure}}
    \centering
    \includegraphics[scale=0.70]{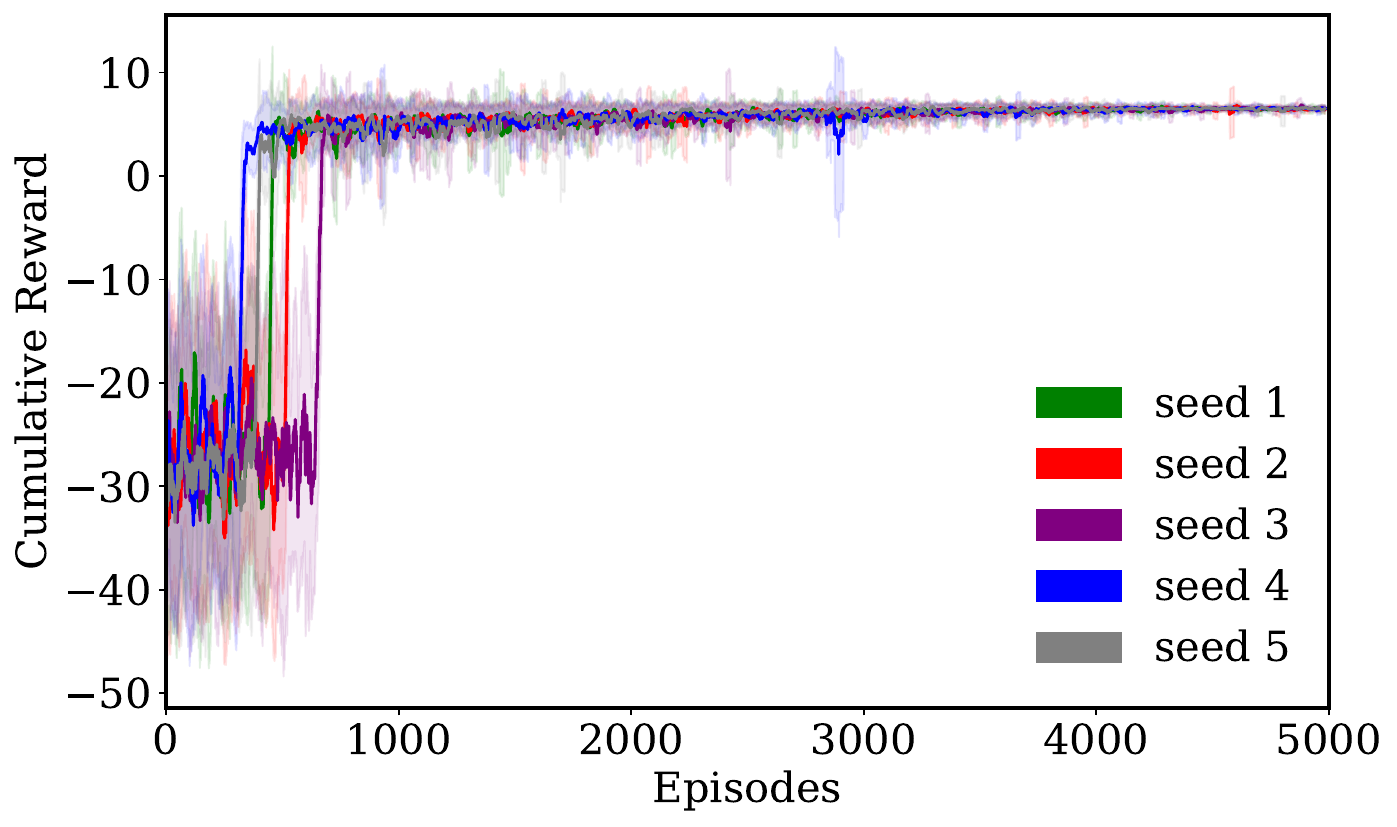} 
    \caption{ The plot depicts that the convergence of RadDQN is not dependent on the choice of seed. This is obtained considering case I (cf. Figure 3; a scenario where two radiation sources of equal strength placed at (2,2) and (7,7)). Five random seeds are chosen to check the sensitivity of training stability and convergence using RadDQN.}
    \label{fig:random_seed}
\end{figure}